\documentclass[conference]{IEEEtran}
\IEEEoverridecommandlockouts

\usepackage{cite}
\usepackage{amsmath,amssymb,amsfonts}
\usepackage{algorithm}
\usepackage{graphicx}
\usepackage{textcomp}
\usepackage{xcolor}
\usepackage{bbm}
\usepackage{url}
\usepackage{cite}
\usepackage{balance}
\usepackage{float}
\usepackage{graphicx}
\usepackage{subcaption}

\ifCLASSINFOpdf
\else
\fi
\DeclareMathAlphabet\mathbfcal{OMS}{cmsy}{b}{n}
\usepackage{algorithm}      
\usepackage{algorithmic} 
\usepackage{varwidth}% http://ctan.org/pkg/varwidth
\begin{document}

\title{Context-Aware Target Classification with Hybrid Gaussian Process prediction for Cooperative Vehicle Safety systems}
% Hybrid-learning-based prediction for robust context aware cooperative vehicular safety systems
% Prediction based target classification using a novel Gaussian process regression method for robust CVS systems
% ***prediction based target classification using a novel Gaussian process regression method for robust CVS systems***
% Context-aware target classification module with hybrid-learning-based prediction for CVS systems
% Model and Context Aware for adaptive and robust cooperative vehicular systems

\author{Rodolfo Valiente$^*$, Arash Raftari$^*$, Hossein Nourkhiz Mahjoub$^*$\\ 
Mahdi Razzaghpour$^*$, Syed K. Mahmud$^{**}$, Yaser P. Fallah$^*$ \\
$^*$Connected and Autonomous Vehicle Research Lab  (CAVREL) \\ Department of Electrical and Computer Engineering, University of Central Florida, Orlando, FL, USA\\
$^{**}$Hyundai America Technical Center, Inc. (HATCI) \\ Superior Township, MI, USA
\thanks{This material is based upon work partially supported by the National Science Foundation under Grant No. CNS-1932037, NSF CAREER Grant 1664968 and in part by the Hyundai America Technical Center, Inc. (HATCI).}
}

\maketitle

\begin{abstract}
Vehicle-to-Everything (V2X) communication has been proposed as a potential solution to improve the robustness and safety of autonomous vehicles by improving coordination and removing the barrier of non-line-of-sight sensing. Cooperative Vehicle Safety (CVS) applications are tightly dependent on the reliability of the underneath data system, which can suffer from loss of information
%However, it is still challenging to attain the full benefits of this Cooperative Vehicular Safety (CVS) systems 
due to the inherent issues of their different components, such as sensors’ failures or the poor performance of V2X technologies under dense communication channel load. Particularly, information loss affects the target classification module and, subsequently, the safety application performance. 
To enable reliable and robust CVS systems that mitigate the effect of information loss, we proposed a Context-Aware Target Classification (CA-TC) module coupled with a hybrid learning-based predictive modeling technique for CVS systems. The CA-TC consists of two modules: A Context-Aware Map (CAM), and a Hybrid Gaussian Process (HGP) prediction system. Consequently, the vehicle safety applications use the information from the CA-TC, making them more robust and reliable. The CAM leverages vehicles’ path history, road geometry, tracking, and prediction; and the HGP is utilized to provide accurate vehicles' trajectory predictions to compensate for data loss (due to communication congestion) or sensor measurements' inaccuracies. Based on offline real-world data, we learn a finite bank of driver models that represent the joint dynamics of the vehicle and the drivers' behavior. We combine offline training and online model updates with on-the-fly forecasting to account for new possible driver behaviors. Finally, our framework is validated using simulation and realistic driving scenarios to confirm its potential in enhancing the robustness and reliability of CVS systems.
\end{abstract}


% Vehicle-to-Everything
% Cooperative Vehicle Safety
% Context-Aware Target Classification
% Context-Aware Map 
% Hybrid Gaussian Process
% Connected and Autonomous Vehicles
% Basic Safety Messages
% Remote Vehicles
% Host Vehicle
% Dedicated Short-Range Communication
% Target Classification
% Remote Vehicle Emulator
% Wireless Safety Unit
% Vehicle-to-Vehicle
% Forward Collision Warning
% convolutional neural network
% Kalman filters
% Constant Speed
% Constant Acceleration
% Long Short-Term Memory
% Auto Encoder
% Gaussian Process

\section{Introduction}
\begin{figure}[t]
 \centering
 \includegraphics[width=.48\textwidth]{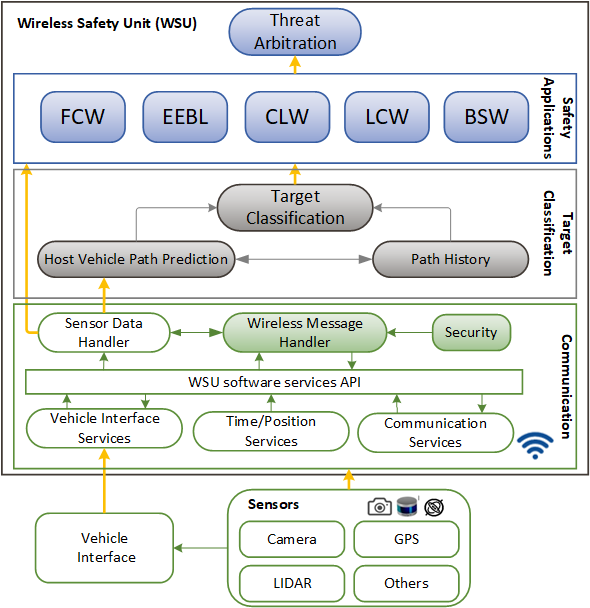}
     \caption{Cooperative Vehicle Safety system block diagram.}
     \label{Figure:CVS}
 \end{figure}

In a Cooperative Vehicle Safety (CVS) system, Connected and Autonomous Vehicles (CAV) take advantage of the data acquired from communication to extend their situational awareness and improve safety~\cite{zhou2022aicp,miucic2018connected,golestan2016situation}. In CVS systems, Remote Vehicles (RV) frequently broadcast their state information to the neighboring vehicles over a wireless channel in the form of an information message, e.g, Basic Safety Messages (BSM)~\cite{Rezaei2010,Zhang2017a} using Vehicle-to-Everything (V2X) communication. V2X communication allows the information from RVs to be available at the Host Vehicle (HV) over the wireless network extending the situational awareness of CAVs and alleviating the limitations of the sensor-based systems that rely on line-of-sight sensing~\cite{valiente2019controlling, razzaghpour2021impact, valiente2020dynamic}. At the HV, the information messages are processed to monitor the neighboring road participants and create a real-time map of all objects in its vicinity. The crash warning algorithms and other safety applications access the real-time map regularly to identify potentially dangerous situations. Therefore, the V2X communication technologies have a significant impact on the accuracy of safety applications. In that sense, two major V2X technologies, Cellular-V2X (C-V2X) and Dedicated Short-Range Communication (DSRC) have been developed~\cite{Shah2019,toghi2018multiple}. Nevertheless, they suffer from information loss and have scalability limitations in real-time applications affecting the quality of the real-time map and the CVS system's performance~\cite{Fallah2015,Fallah2016b,Rezaei2010,Zhang2017a}.
CVS architectures separate the design of applications from the communication system allowing flexibility and simpler adoption of different communication technologies. Figure ~\ref{Figure:CVS} shows a CVS system that consists of a communication and sensor layer, a Target Classification (TC) layer, and a safety application layer~\cite{ahmed2011vehicle,thorn2018framework}. TC classifies the neighboring RVs based on their relative locations with respect to the HV. While the safety applications have been extensively studied~\cite{ahmed2011vehicle,thorn2018framework,valiente2020dynamic,misener2005cooperative,lyu2019characterizing}, there are limited studies on TC~\cite{Lee2005,Kim2016}. Current TC solutions are limited in their context-awareness and prediction capabilities, making them vulnerable to communication losses, subsequently affecting the performance of the safety applications~\cite{Lee2005,Kim2016}.
%The main focus of this research is addressing the performance degradation challenges in CVS safety applications due to imperfect information. The imperfection exists in both on-board sensor and V2X communication pipelines, because of onboard sensors’ failures and V2X performance degradation under heavy communication traffic loads (congested scenarios). 
%Future trajectories of surrounding objects are the most crucial information for the cooperative vehicular safety and comfort applications

To address this challenge we present a Context-Aware Target Classification (CA-TC) module for mitigating the effect of communication loss or sensor failure. The CA-TC fits within the CVS system block and consists of two modules: a Context-Aware Map (CAM) and a Hybrid Gaussian Process (HGP) prediction system as shown in Figure~\ref{Figure:TC}. The CAM module separates application and information/perception subsystems and leverages vehicles’ path history, road geometry, tracking, and prediction. The prediction module is a non-parametric Bayesian inference modeling scheme, Gaussian Process regression, and is responsible for vehicle's trajectory predictions when data is not received due to communication losses or sensor failure. The prediction system allows the CA-TC to precisely predict future behaviors and compensate for information loss. In this paper, we focus on improving the TC layer in Figure~\ref{Figure:CVS}. While we particularly proposed a suitable prediction method for our architecture, the CA-TC allows the use of any prediction method to address the aforementioned challenge.

% TODO: more explanation can be added about the fig

% Important
%Our prediction approach is robust to communication loss, no-parametric, allow on-the fly learning and prediction,  and meet the real time computational requirements., capture the unforeseen patterrsn on the data on-the-fly, is receiver side solution, hybrid HGP , clusterd reduced kernel bank and on the fly prediction and new kernel generation

%It is shown that our proposed prediction scheme has superior performance in terms of PTE and application WA when compared with other approaches. We compare the performance of different prediction schemes (baselines) based on those metrics. 

Our main contributions are as follows:
% Our main contributions are:
\begin{itemize}
  \item We proposed a Context-Aware Target Classification module for CVS systems that alleviate the effect of information loss on the performance of safety applications.
 \item A novel Hybrid-learning-based Gaussian Process prediction method is presented. It relies on a bank of driver models learned from real data in an offline manner to perform forecasting on-the-fly, while also allowing online updates of the driver models, accounting for new possible driving behavior.
 \item The proposed system is evaluated in a simulation and hardware-in-the-loop utilizing a Remote Vehicle Emulator (RVE) based on a DSRC Wireless Safety Unit (WSU), which facilities the analysis of the CVS applications and the communication system performance. Additionally, we study the impact of using different prediction techniques and demonstrate the performance gains caused by the proposed prediction method.
%  The proposed system is validated with comprehensive simulations and in a real environment using a Remote Vehicle Emulator (RVE) based on a DSRC Wireless Safety Unit (WSU), which allows the joint study of the CVS applications and its underlying communication system. Additionally, we study the impact of using different prediction techniques and demonstrate the performance improvements of the proposed prediction method.
\end{itemize}

\section{Literature Review}

\subsection{Cooperative Vehicle Safety Systems}
CVS systems are an active area of research in vehicle safety~\cite{Sengupta2007,Jami2022AugmentedDB,Kim2016,ye2019evaluating}. In~\cite{Forkenbrock2009,Jamson2008} the information from V2X communication is used for increased situational awareness and improve safety, and authors in~\cite{Lee2005,Kiefer2003} present a CVS system that continuously tracks all RVs in the vicinity of the HV. CVS systems rely on the accuracy of the real-time map of RVs which is processed by the safety applications to identify potential threats. The architecture must be modular and flexible enough to allow variations in communication rate and sensor limitations~\cite{caveney2010cooperative}.
% CVS systems are an active area of research. CVS applications use the information obtained from wireless communication or sensory systems for situational awareness~\cite{Forkenbrock2009,Jamson2008} and tracking vehicles in the vicinity of the HV~\cite{Lee2005,Kiefer2003}. Therefore, most of the CVS systems work based on creating and analyzing a real-time map of the vicinity. The map is continuously analyzed by applications to detect possible hazards. The first task in any CVS system is to detect the possibility of collision and then act to avoid it either through warning the driver or taking evasive maneuvers. The architecture must be modular and flexible enough to allow variations in communication rate and sensor limitations~\cite{caveney2010cooperative}. 

Figure~\ref{Figure:CVS} presents a CVS system architecture~\cite{ahmed2011vehicle,thorn2018framework}.
% It consists mainly of a communication and sensors layer, a Target Classification (TC) layer, and an application layer. 
The modules include the services modules, WSU software API, sensors and wireless data handler, the TC, and safety applications. The WSU software services modules are generic modules supplied by the WSU that provide services and an API to enable applications to interface to the CAN bus, GPS receiver, and wireless. Based on the data from surrounding vehicles, the TC classifies the locations of the RVs within a specified radius of the HV and provides relative position and velocity metrics for the classified RVs.
% The TC process the data from the surrounding vehicles classifies the locations of communicating RVs within a specified radius of the HV and provides relative position and velocity metrics for the classified RVs. 
The application modules consist of multiple Vehicle-to-Vehicle (V2Vs) applications and evaluate potential safety threats based on inputs from the system framework modules~\cite{ahmed2011vehicle}. 
%The arrows and text between the modules show the primary data flow.

\subsection{Cooperative Vehicle Safety Applications}
% Traditional safety applications rely on ranging sensors such as radar, LiDAR, or other sensors to detect vehicles in front. In cooperative perception systems, the information available through V2V communication can be used to detect the presence and position of other vehicles on the road, this information is sent by RVs over a wireless channel (in the form of BSMs). 
Different from non-cooperative safety applications that depend on the line-of-sight sensors, cooperative perception systems leverage the information accessible through V2V communication to identify the presence and location of the RVs increasing the perception range.
In the vehicle safety communication report from NHTSA~\cite{ahmed2011vehicle,thorn2018framework} the main cooperative safety applications within CVS were described, such as Cooperative Forward Collision Warning (FCW), Control Loss Warning (CLW), Emergency Electronic Brake Lights (EEBL), Blind Spot Warning (BSW), Lane Change Warning (LCW) among others.
%Traffic Signal Violation Warning, Control Loss Warning (CLW), Curve Speed Warning, Emergency Electronic Brake Lights (EEBL), Cooperative Forward Collision Warning (FCW), Left Turn Assistant, Lane Change Warning (LCW), Stop Sign Movement Assistance, Do-Not-Pass Warning (DNPW), Blind Spot Warning (BSW) among others. Safety applications have been extensively studied. 
In~\cite{tahmasbi2017implementation} a Cooperative Vehicle-to-Pedestrian safety application is implemented to provide extended situational awareness and threat detection, reducing crashes and improving safety. Authors in~\cite{sepulcre2013cooperative} present a cooperative testing platform to evaluate the efficacy of CVS applications under challenging conditions and a Collision Warning Avoidance (CWA) system is presented and evaluated in~\cite{lee2005evaluation}.
%While the safety applications have been widely explored the TC layer has received less attention. However, it is a main component in the CVS architecture and the performance of all the applications rely on an accurate and robust TC \cite{Lee2005}, \cite{Kim2016}. 
%Cooperative forward collision warning system is designed to aid the driver in avoiding or mitigating collisions with the rear-end of vehicles in the forward path of travel through driver notification or warning of the impending collision. The system does not attempt to control the host vehicle in order to avoid an impending collision. A cooperative forward collision warning system would use information communicated from neighboring vehicles via vehicle-to-vehicle communication in addition to forward looking sensor data to address these shortcomings.

\subsection{Target Classification (TC)}
In order to realize the above-mentioned safety applications, the surrounding vehicles should be tracked and classified. While the safety applications have been widely explored, the TC layer has received less attention~\cite{Lee2005, Kim2016}. However, it is a main component in the CVS architecture and the performance of all the applications relies on an accurate and robust TC~\cite{Lee2005, Kim2016}. 
%However, CVS applications rely on communication this information is uncertain and vulnerable to communication losses due to network congestion and errors caused by sensor limitations. In fact, one of the main challenges of V2V communication is the issue of channel congestion and communications losses\cite{Sepulcre2013,Ahmed-Zaid2011}. Moreover, in current approaches, design of safety applications is integrated with information system. This leads to difficulties in adopting safety applications to new and emerging sensing and communication technologies. 
% The majority of existing solutions keep applications entirely dependent on the information system, without a context-aware TC layer, which renders it difficult to handle data fusion and process the uncertain information \cite{Sengupta2007}. 
In that direction, a machine learning-based TC for Millimeter-Wave Radar is proposed in~\cite{cai2021machine}, it uses an artificial neural network and convolutional neural network (CNN) and shows acceptable performance in a specific domain. In~\cite{Kim2016} a TC layer is proposed to classify the RVs with respect to the HV, which considers road geometry, the local map, the path history, and a simple vehicle tracking algorithm based on a Kalman filter. These approaches are limited in their prediction capabilities making them vulnerable to communication losses and sensor failures. In order to mitigate these issues, we propose a CA-TC layer and a robust and suitable prediction approach for the TC layer.

\subsection{Tracking and Prediction}
% Most of the CVS applications rely on the ideal communication for the performance of the safety applications, making them vulnerable to communication losses due to network congestion and errors caused by sensor limitations.
% As discussed, one of the main challenges of V2V communication is the issue of channel congestion and communications losses~\cite{Sepulcre2013,Ahmed-Zaid2011}.
%To address the performance degradation challenge in CVS safety applications due to imperfect information tracking and prediction approaches have been proposed. 
In order to overcome CVS performance degradation caused by non-ideal communication, methods for tracking and predicting information have been proposed~\cite{Parker2007,Baek2017,Painter1990,Chu2010,hnmahjoub:syscon19}.
%In~\cite{Parker2007}, a constant position estimator has been proposed; however, this stationary approach is limited, especially in congested scenarios where Packet Error Rate (PER) is high,  the position prediction errors increase with speed and the performance of safety applications detrimentally decrease. 
% In industrial usage by auto manufacturers, the most widely granted assumption for position prediction is the constant speed (CS) or acceleration (CA) model. The underlying reason behind these assumptions is the standardization of the BSM transmission rate to 10Hz. Given the short time duration between each data packet and the assurance of regular availability of BSMs from neighborhood vehicles, this is a rational assumption for an optimized ideal domain. However, in a practical scenario, the assumption loses legitimacy because of the limitation of communication technology. Especially in highly congested urban environments, there is always a high possibility of loss or delay of packet reception due to collision~\cite{yfallah:mbcsyscon}. 
The constant speed or acceleration model is the most frequently accepted assumption for prediction by conventional vehicle manufacturers. The fundamental rationale for using those models is the standardization of the BSM communication rate to 10Hz, given the short time between BSM the models remain valid under ideal communication channels. Nevertheless, because of the limitations of communication technology, the assumption is not always valid in real scenarios, particularly in heavily congested situations~\cite{yfallah:mbcsyscon}.
%For a longer than usual delay between two consecutive packets, the prediction might be highly erroneous from the CS assumption. For the CA assumption, a long delay in message reception can diverge and push the predicted speed to infinity~\cite{yfallah:mbcsyscon}. 

% Kalman filters (KF) have been also used widely for tracking position and velocity~\cite{Baek2017,Painter1990}. This is likely due to the relative simplicity and robust nature of the filter itself. 
% It has a similar architecture as the deterministic alpha-beta filter, with a difference in implementations that relies on the method for designing the adjustable “tuning” coefficients of the filters \cite{Baek2017}. In \cite{Painter1990}, a joint Kalman and alpha-beta design is presented, showing how they are related, and comparing them on a common basis. However, the qualitative analysis of the results including comparison with other approaches is not provided.
% Chu et al.~\cite{Chu2010} propose a vehicle velocity prediction based on adaptive KF. This method is evaluated under a variety of driving conditions and the prediction values are compared with simulator values from CarSim. It was proved that the proposed method is robust and can improve the prediction accuracy of velocity.
%In this paper we perform prediction of RV position when messages are lost, and we consider that each vehicle has precise information about its own states (HV and RV own information is assumed to be correct). 
Kalman filters have been also used widely for tracking position and velocity~\cite{Baek2017,Painter1990}. This is likely due to the relative simplicity and robust nature of the filter itself.  In~\cite{Chu2010} is presented an adaptive KF to predict vehicle velocity, the approach is tested in a variety of driving scenarios showing a good performance in velocity prediction.

% NEW REVIEW
More recent effective approaches for trajectory prediction include recurrent neural networks ~\cite{morton2016analysis, alahi2017learning}, gaussian mixture models~\cite{das2010block, wang2007gaussian} and Long Short-Term Memory NNs~\cite{guan2019intelligent, altche2017lstm}. In \cite{guan2019intelligent} is described an LSTM model to predict vehicle trajectories, leveraging the LSTM network to infer the temporal relationship from previous sequence data.
Authors in \cite{guan2019intelligent} propose a coupling LSTM model to effectively predict the future trajectories of vehicles. The LSTM network can learn the temporal relation from the historical sequence data and is employed as the basic prediction model for all vehicle trajectories with common parameters.
Following the recent developments in that field of generative models, several works make use of autoencoder-based solutions~\cite{goodfellow2014generative, sohn2015learning}, proposing solutions that include Recurrent Variational Autoencoder, Conditional Variational Autoencoder or Generative Adversarial Network ~\cite{goodfellow2014generative, sohn2015learning}. Autoencoders have been used effectively to learn a better representation and for a prediction task in previous works~\cite{valiente2022prediction,toghi2022social,toghi2021towards}. In ~\cite{valiente2022prediction} a predictive autoencoder is proposed to produce future observation, predicting velocity map images directly and kinematics predictions using a the predictive autoencoder. Authors in ~\cite{ neumeier2021variational } present a variational autoencoder used for  prediction with partial interpretability. The proposed model achieves good performance, however, the expensive training cannot be done on-the-fly while new data is available, which makes it harder to learn from current information and adapt to new scenarios.    

% NEW REVIEW

To address those challenges we propose a novel hybrid learning approach based on Bayesian inference.
% Different Bayesian inference strategies have been explored in previous works in the context of model-based communication (MBC), with promising results~\cite{Mahjoub2019b}. In general,
% Non-parametric Bayesian methods allow us to infer the underlying patterns of observed time-series and capture joint driver/vehicle dynamics without imposing any presumptions or limitations on their characteristics. This phenomenal aspect relaxes the learning process from being bound to specific function patterns. In other words, the complexity of a model which is derived in a non-parametric Bayesian inference framework is automatically adapted to the observed data; and hence can both avoid creating over-complex models and capture the unforeseen patterns in the data on-the-fly.
By leveraging non-parametric Bayesian approaches we can predict fundamental patterns of observable time series and capture coupled driver/vehicle dynamics without enforcing any assumptions on the model parameters. This extraordinary feature frees the learning process from the constraints of certain function patterns. The complexity of a model created in a non-parametric Bayesian inference framework is automatically adjusted to the observed data, allowing it to avoid over-complex models while still capturing unforeseen patterns in the data on-the-fly.
% Gaussian Process (GP), which is one of the most powerful non-parametric Bayesian inference methods, has shown an enticing performance improvement in our recent studies in terms of required packet generation rate and also position tracking accuracy under network congestion~\cite{hnmahjoub:syscon19}. GP tries to regress the observed time-series realizations by putting a prior distribution directly over the function space, instead of function parameters space, in a form that any finite subset of draws from this distribution represents a multivariate Gaussian random vector~\cite{Rasmussen2003}. This method by nature adapts the model complexity to the observed data and makes it capable of capturing different trends while they appear in training data.
Particularly, within the non-parametric Bayesian inference approaches, Gaussian Process (GP) has demonstrated a significant performance improvement in terms of packet generation rate as well as position tracking accuracy under network congestion~\cite{hnmahjoub:syscon19}. GP attempts to regress observed time-series realizations by projecting a prior distribution directly over the function space, rather than the function parameters space, in such a way that any finite subset of draws from this distribution represents a multivariate Gaussian random vector~\cite{Rasmussen2003}. This approach adjusts the model complexity to the observed data, allowing it to capture distinct patterns as they arise during training. This idea has been utilized in \cite{Mosh2111:Gaussian,Safety_ECC} to enhance the performance of Cooperative Adaptive Cruise Control (CACC) in congested traffic scenarios.
% Previous works have demonstrated that within the  Model-Based Communication (MBC) context, a modeling scheme that employs GP as its inference method has a notable superiority over the other modeling approaches \cite{hnmahjoub:syscon19,Mahjoub}. However, previous papers present a transmitter-side solution that uses GP for prediction using a MBC paradigm which requires fundamental changes in current vehicular data dissemination standards and its data format. Different from that, in this paper, we proposed a receiver-side solution using GP to predict future states of the RV based on the history which can be utilized with current vehicular network standards and BSM data format.

Previous works have demonstrated that within the context of Model-Based Communication (MBC), a modeling scheme that uses GP as its inference technique outperforms other modeling approaches~\cite{hnmahjoub:syscon19,Mahjoub}. Different from other works that provide a transmitter-side approach using GP for prediction via an MBC paradigm we propose a receiver-side approach that leverages GP to forecast the trajectories of the RV, therefore, the proposed architecture does not require modifications in the current V2X standards.

% LSTM and other ML proposed model achieves good performance, however, is not able to adapt to new scenarios and expensive training cannot be done offline while new data is available.  

% new pipeline and algorithms what we proposed for cooperative position prediction based on BSMs. Using GPR for estimating a quantity is of course not new, however, it is not possible to directly apply GPR to a problem. We should think of GPR as a mathematical tool. What we are proposing is a method of learning behavior of a vehicle (trajectory) , learned from speed and heading, then using a novel algorithm to decide how long a particular learned model remains valid (given tunable hyper-parameters), before new models are trained. Also, we are proposing to use a Kernel bank (model bank), a set of many kernel parameters that represent many GPs that can be applicable to a vehicle behavior, and then using the bank among vehicles. The bank will be used by each vehicle to inform other vehicles (for example using V2V) to predict its state. Our prediction (estimation) method also utilizes a check between a deterministic physical model (constant velocity) and current selected GPs to decide whether a GP is to be used for prediction.
% Our method looks at how a RV future position (based on speed and heading) can be predicted using a bank of models (obtained based on our specific model generation algorithm) and a new model generated on-the-fly (derived based on our proposed algorithm)

\section{Context-Aware Target Classification (CA-TC)}
%Our context-aware target classification fits within the CVS system block and consists of two modules. A context-aware dynamic map (CAM) and a hybrid-learning-based GP prediction system.  

\begin{figure*}[t]
 \centering
 \includegraphics[width=.9\textwidth]{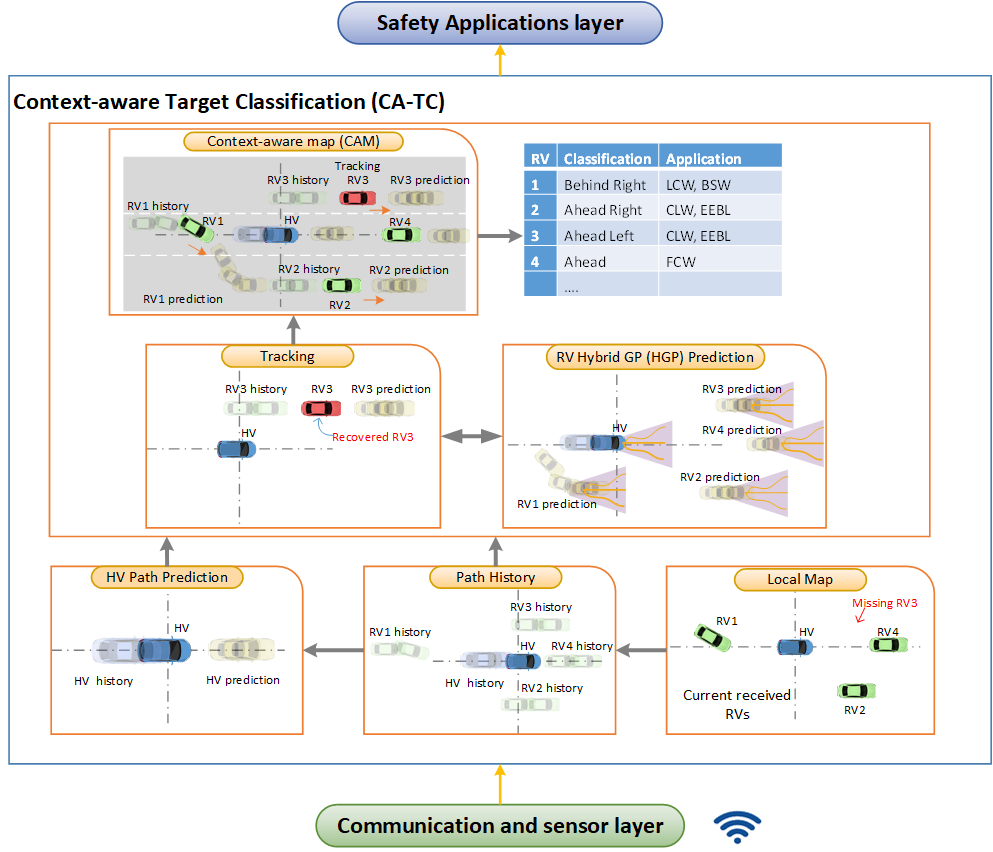}
     \caption{Context-Aware Target Classification module with HGP prediction for CVS systems. The HGP prediction system is responsible for accurate vehicle predictions and the CAM leverages vehicles’ path history, road geometry, tracking, and prediction to improve the robustness and reliability of the safety applications. In the figure, HV is in blue, RVs are in green, a missing RV is in red, predictions are in shadow yellow, and the vehicle's history is in shadow green.}
     \label{Figure:TC}
 \end{figure*}

In this section, we present the CA-TC. The CA-TC fits within the CVS system block in Figure~\ref{Figure:CVS}. The proposed CA-TC is presented in Figure~\ref{Figure:TC} and consists of the HV path prediction, the path history of all vehicles, the current received BSMs from RVs (Local Map), a tracking module, and the HGP system, all of which are integrated into the CAM, in which data fusion, history, and predictions are used for lane estimation and local map reconstruction creating a real-time context map of the surroundings. 
% A Context-Aware Map (CAM) and a hybrid-learning-based GP (HGP) prediction system. The general architecture is presented in Figure~\ref{Figure:TC}.
By leveraging tracking and prediction, CA-TC provides accurate information that improves the performance of safety applications. The CA-TC enables the applications to work based on the CAM rather than individual BSMs or sensory data. Using the CAM, the RVs are classified based on the relative current and predicted location with respect to HV.
%both estimation and classification of RVs locations which is used by applications to determine collision hazards in real-time, improving consequently the classification accuracy of the applications. 
Therefore, it allows the safety applications to be triggered not just by the local map and history but also based on the predictions, either when packages are lost or when the prediction system detects a possible warning. For instance, as an illustration in Figure~\ref{Figure:TC}, the local map is showing RV1 as behind, but the CAM is predicting RV1 will be behind right, so RV1 will be also processed by LCW and BSW safety application to check possible future hazards based on the CA-TC information. The proposed CA-TC allows applications to be independent of underlying communication and sensing systems, and the architecture is adaptable to new sensing technologies.
% By using the proposed CA-TC the applications are agnostic to underlying communication technology and sensors and the design is flexible in employing new communication or sensing technologies. 
% Safe apliations will be trigger by local map and prediction.... As e.g Fig ....
%With the proposed DOM, safety applications work based on the dynamic map rather than individual BSMs, increasing robustness to communication failures by improving the performance of applications in presence of outdated information and loss of messages (e.g. in congestion scenarios and lossy channels). The proposed DOM made the design of an application independent of how and when BSMs are received and processed, allowing flexibility to add new applications, new communication technologies (any communication module can be used, e.g. DSRC, C-V2X, mmWave \cite{Niu2015}) and new sensors, and consequently, facilitating sensor fusion and cooperative perception. The resulting dynamic map have higher fidelity than current state of the art methods.

In the CA-TC (Figure~\ref{Figure:TC}), the map-update module (Local Map) is in charge of refreshing the real-time map with the latest received information from sensors, GPS, and communication module. Thus, the real-time map database keeps the records of its surrounding entities, such as all neighboring vehicles, and other detected objects. Every local map record consists of the latest available information about a particular neighboring entity, such as its position information, speed, size, etc. The local map is created and updated based on the latest HV and RVs position. The path history (PH) module keeps track of HV and RVs history. It keeps a buffer of the vehicle's most recent position and sensor data points and computes concise representations of the vehicle's actual PH based on the allowable position error tolerance between the actual vehicle path and its concise representation, and updates the PH concise representation periodically for use by the other modules. The HV path prediction module makes use of the HGP prediction algorithm to forecast the future path of the HV. The tracking module utilizes the HGP prediction algorithm together with the previous and current local map information to correct the vehicle's position, recover current missed RVs and filter possible outliers. The HGP prediction system uses the information from PH, tracking, and current local map records for future path prediction points for all vehicles. By having an accurate RV prediction, the CA-TC allows safety applications to be triggered based on forecasting and not just current received information.
%In the map, GPS information is converted into Earth-Centered, Earth-Fixed format (ECEF). The ECEF format is offset to the center of the vehicle on the ground and then transformed into East-North-Up (ENU) coordinates. The tracking module utilizes a GP together with the previous and current map information to correct the vehicle's position, recover current missed RVs and filter possible outliers.
% GP-based tracking and prediction...

All the aforementioned modules are used by the CAM for an extended situational awareness subsystem that is responsible for delivering the latest and most accurate information to the safety application to identify present and future possible threats and generate appropriate notifications and signals. While the CAM can leverage information from local sensors and V2X communication, in this work, we focus on the communication side and study the impact of communication imperfections on CVS applications.
% While the CAM considers both the local perception map and cooperative map. In which the local perception map process the received data from the HV local sensors (vehicles sensor such as camera, LIDAR, etc.), and the cooperative perception map process the data received through wireless communication. We focus on the communication side and study the impact of communication imperfections on the CVS applications.
%While the CAM allows to integrate the local and RV perception map into a surrounding dynamic map for HV.  Additionally, it helps to alleviate both imperfections in communication and perception systems, in this paper our results focus more in the communication losses. 
%In this paper we propose an architecture that separate application (e.g. Forward Collision Warning (FCW)) from information/perception subsystems (sensors and communication modules). Our architecture introduces a dynamic middle layer called Dynamic Object Map (DOM), Figure \ref{Figure:DOM} shows a simplified representation of a DOM-based architecture. 

\subsection{Target Classification}

\begin{figure}[t]
 \centering
 \includegraphics[width=.4\textwidth]{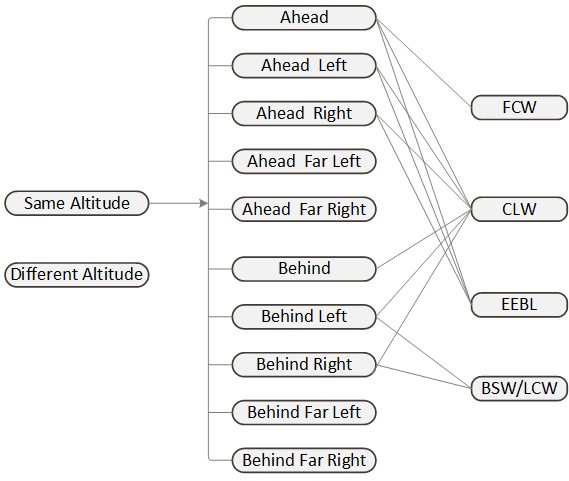}
     \caption{Target Classifications areas used by the safety applications.}
     \label{Figure:TC_apps}
 \end{figure}

\begin{figure}[t]
 \centering
 \includegraphics[width=.48\textwidth]{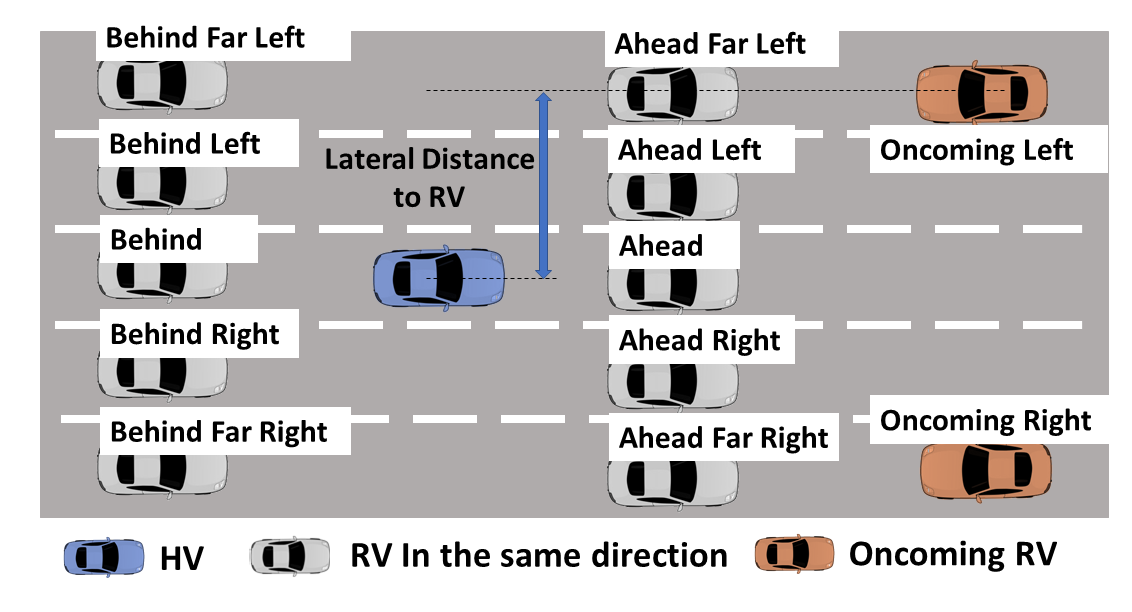}
     \caption{Target Classifications of the RVs relative to the HV.}
     \label{Figure:TC_class}
 \end{figure}

With the information provided by the CAM, the RVs are classified in different areas based on the relative locations with respect to the HV. In addition to the classification of RVs, the CA-TC also provides the vehicles' predictions, relative speeds, headings, lateral and longitudinal offsets, and tracking positions. Figure~\ref{Figure:TC_class} shows the classification of the RVs relative to the HV. The CA-TC computes the RVs classifications area based on the longitudinal and lateral distances with respect to HV~\cite{Lee2005,Kim2016}.
based on the relatives' longitudinal distance the RVs are classified as,
\begin{equation}
\label{equ:tc_longi}
RV_{class} = 
\begin{cases}
Xrel\geq  0, &\quad \textrm{Ahead} \\
Xrel <  0, &\quad \textrm{Behind}  \\
\end{cases}
\end{equation}
Next the longitudinal distance classifications are combined with the relatives' lateral distance ($ld$) information to classify the RVs as,
\begin{equation}
\label{equ:tc_lateral}
RV_{class} = 
\begin{cases}
ld \in  (1.5w_{lane}, \infty), &\quad \textrm{Far Left} \\
ld \in (0.5w_{lane}, 1.5w_{lane}], &\quad \textrm{Left}  \\
ld \in  [-0.5w_{lane}, 0.5w_{lane}], &\quad \textrm{On Centre}   \\
ld \in  [-1.5w_{lane}, -0.5w_{lane}), &\quad \textrm{Right}  \\
ld \in  (-\infty , -1.5w_{lane}),  &\quad \textrm{Far Right}  \\
\end{cases}
\end{equation}
in which $w_{lane}$ is the width of the lane. Finally, for the oncoming/ongoing traffic, the lane heading angle $\phi_{lane}$ is compared with the RVs heading angle $\phi_{RV}$ to classify the RVs as,
\begin{equation}
\label{equ:tc_heading}
RV_{class} = 
\begin{cases}
|\phi_{RV}  - \phi_{lane}| \leq \Delta\phi_{ongoing},  &\quad \textrm{Ongoing}  \\
|\phi_{RV}  - \phi_{lane}| \geq  \Delta\phi_{oncoming},  &\quad \textrm{Oncoming}  \\
\textrm{else, } &\quad \textrm{not classified}
\end{cases}
\end{equation}
where ongoing vehicles are vehicles in the same direction, and oncoming vehicles are vehicles in the opposite direction. $\Delta\phi_{ongoing}$ and $\Delta\phi_{oncoming}$ are the threshold values for classification. Based on these RV classifications the safety applications are evaluated. Figure~\ref{Figure:TC_apps} depicts the mapping of RV classification to various safety applications. Based on the data given by the CA-TC, the safety application modules independently analyze possible safety hazards. This demonstrates that appropriate TC is crucial for the triggering and functioning of safety applications, as the warnings are dependent on the RVs being classified correctly. For example, in the event of an approaching rear-end collision with a vehicle ahead in the same lane and direction of travel, the TC will classify the vehicle as Ahead and the FCW application will be evaluated as indicated in Figure~\ref{Figure:TC_apps}.

\section{Hybrid-learning-based Gaussian Process (HGP) prediction system}
%We proposed a GP inference method that finds a mathematical forecasting model-based on the observed history of RVs. This model is a more powerful solution to address the congestion challenge in dense driving scenarios.

The prediction system is used at the HV (receiver side) to regress the observed time-series realizations from PH, capturing the distinct patterns as they emerge in the data. Each observation point is chosen from Gaussian random variables that are not independent and have temporal relationships with their predecessors and successors. The correlation describes the temporal relationship between observations, making GP a useful tool for detecting patterns in time series.
% The prediction system is used at the receiver side to regress the observed time-series realizations from PH. This method by nature adapts the model complexity to the observed data and makes it capable of capturing different trends while they appear in the data. Different patterns are essentially recognized through GP. More precisely, each observation point is assumed to be drawn from a Gaussian random variable. However, these normal random variables are not independent and have temporal correlations with their predecessors and successors. This correlation models the temporal relation between observations and makes GP a powerful method to capture different patterns within time series.

% The set of $m$ observed values is modeled as an m-dimensional multivariate Gaussian random vector which could be defined using a mean vector of length $m$ and an $m \times m$ covariance matrix. This covariance matrix, or GP kernel, is the core asset by which GP recognizes the inherent behavior of time series through the observed time series history and forecasts its future. The core GP components could be mathematically formulated as follows. 
The set of $m$ observed values is represented by an m-dimensional multivariate Gaussian random vector, which is defined by an $m \times m$ covariance matrix and a $m$ mean vector. This covariance matrix, often known as the GP kernel, is the foundation upon which GP detects and anticipates the underlying behavior of time series based on their recorded history. The fundamental GP components can be expressed mathematically as follows,
%For more details one can refer to [26].
\begin{multline}
\label{gp-generaleq}
\hspace{-5pt} f(t) \sim gp \big(m(t), k(t,t')\big), \hspace{137pt}\\
\hspace{-2pt}\{X_{i}\}_{i=1,2,...,m} = \{f(t_i)\}_{i=1,2,...,m} \sim \mathcal{N}(\mu,\,\Sigma), \hspace{10pt}\\
\hspace{1pt}\mu = \big[m(t_1), ..., m(t_m)\big]^T, \hspace{70pt}\\
\hspace{3pt}\Sigma_{ij} = \kappa(t_i, t_j) \ \forall i,j \in \{1,2, ..., m\},\hspace{30pt}
\end{multline}
where $X_{i}$, $f(t)$, $m(.)$, and $\kappa(.,.)$ are the samples of the vehicles' state, observed or to be predicted at the time $t_i$, the unknown underlying function that the vehicles' states are sampled from, the mean and the covariance functions, respectively.
% \begin{equation}
%   f(t) \sim gp (m(t), k(t,t'))
% \end{equation}

% \begin{equation}
%   \{X_{i}^{Obs}\}_{i=1,2,...,m} = \{f(t_i)\}_{i=1,2,...,m} \sim \mathcal{N}(\overline{\mu},\,\Sigma)\ 
% \end{equation}

% \begin{equation}
%   \overline{\mu} = m(t_i); \ \Sigma_{i,j} = k(t_i, t_j) \ \forall i,j \in \{1,2, ..., m\}
% \end{equation}

%The novel method is divided in two faces: 1-Training, 2-Forecasting. First, training is done offline, the SPMD dataset is used for training and extract a group of significant models that represent the data. 
% In the proposed inference algorithm, instead of working directly on the position time series, the heading and longitudinal speed of the vehicle are treated as two independent time series which should be regressed using GPs. 
Instead of working directly with the position time series, the proposed inference algorithm treats the vehicles' heading and longitudinal speed as two independent time series that are regressed using GPs. The proposed method is divided into two stages: Training and Forecasting. During the offline training procedure, the hyper-parameters of kernel functions are learned based on the observed samples of the vehicles' speed and heading to create a bank of vehicle/driver short-term behavioral models. The well-known realistic data set, Safety Pilot Model Deployment (SPMD)~\cite{SPMD:Data}, is used for training and selecting a group of significant models that represent the data. It was shown through experiments that a limited-size kernel bank suffices to model joint vehicles’ dynamic and driver behavior. Given the availability of this bank of models, during forecasting, the most recent information of RVs and path history is used to select the model with the highest likelihood from the bank and use it to forecast the vehicles' state until new information is received.

\subsection{Training}

During training, the sequences of equally-spaced samples of longitudinal speed and heading, measured every $100 ms$, with size $TW=30$ were used to learn the GP models and subsequently construct the kernel banks. In this work, based on the findings in~\cite{hnmahjoub:syscon19}, a compound kernel of Radial Basis Function (RBF) and a linear kernel is used as the covariance function and the mean function is considered to be zero. Therefore, the covariance function in \eqref{gp-generaleq} can be expressed as:
\begin{equation}
\label{covariance func}
\centering
 \kappa(t,t')=\alpha_{1}^{2} \exp(-\frac{||t-t'||^2}{2\gamma^{2}})+ \alpha_{2}^{2}tt'.
\end{equation}
we used the Leave-One-Out cross-validation to learn a GP model and obtain the set of parameters $\theta=\{\gamma,\alpha_{1}, \alpha_{2}\}$ for each under consideration sequence of speed or heading. Assuming the $i^{th}$ element of the observed speed or heading sequence, $x_i$, is left out, the joint distribution of $x_i$ and the rest of the sequence, $\mathcal{X}_{-i}$, can be expressed as: 
\begin{equation}
\left[\begin{array}{l}
{x}_{i}\\
\mathbf{\mathcal{X}_{-i}}
\end{array}\right] \sim \mathcal{N}\left(\mathbf{0},\left[\begin{array}{ll}
\kappa({t_i},{t_i}) & K_{1}^T\left(t_i, \boldsymbol{t_{-i}}\right) \\
K_{1}\left(t_i, \boldsymbol{t_{-i}}\right) & K_{2}\left(\boldsymbol{t_{-i}}, \boldsymbol{t_{-i}}\right)
\end{array}\right]\right),
\end{equation}
where $K_{1}$ is a $(m-1)\times 1$ vector representing the covariance between $\mathcal{X}_{-i}$ and $x_i$, $K_{2}$ is the $(m-1)\times (m-1)$ covariance matrix of $\mathcal{X}_{-i}$, and $\boldsymbol{t_{-i}}$ represents the time stamps of $\mathcal{X}_{-i}$. $K_{1}$ and $K_{2}$ can be obtained using \eqref{covariance func}.
The conditional probability of $x_i$ given the rest of the sequence can be derived as
\begin{multline}\label{eq:conditional-one}
    \Big(\mathbf{x_{i}} \mid  \mathbf{t}, \mathbf{\mathcal{X}_{-i}}, \mathbf{\theta}\Big) \sim \mathcal{N} (\mu_{i},\sigma_{i}), \\ 
    \mspace{-75mu}\mu_{i}=\mathbf{K_{1}}^{T}[\left(t_{i}, \mathbf{t_{-i}}\right)|\mathbf{\theta}] \mathbf{K_{2}}^{-1}[(t_{-i}, t_{-i})|\mathbf{\theta}] \mathbf{\mathcal{X}_{-i}}, \\
    \mspace{6mu}\sigma_{i}=-\mathbf{K_{1}}^{T}[\left(t_{i}, \mathbf{t_{-i}}\right)|\mathbf{\theta}] \mathbf{K_{2}}^{-1}[(\mathbf{t_{-i}}, \mathbf{t_{-i}})|\mathbf{\theta}] \mathbf{K_{1}}[\left(t_{i}, \mathbf{t_{-i}}\right)|\mathbf{\theta}] \\
    \mspace{18mu}+\kappa[({t_i},{t_i}) \mid \mathbf{\theta})].
\end{multline}
Therefore, the $\log$ probability of observing $x_{i}$ given the rest of the sequence, ($\mathcal{X}_{-i}$), can be calculated as
\begin{equation}
\log p\left(x_{i} \mid \mathbf{t}, \mathcal{X}_{-i}, \mathbf{\theta}\right)=-\frac{1}{2} \log \sigma_{i}^{2}-\frac{\left(x_{i}-\mu_{i}\right)^{2}}{2 \sigma_{i}^{2}}-\frac{1}{2} \log 2 \pi
\end{equation}
Defining the cross-validation objective function as the sum of the $log$ probabilities over all elements of the sequence, i.e., ${L}(\mathbf{t}, \mathcal{X}, \theta)=\sum_{i=1}^{m} \log p\left(x_{i} \mid \mathbf{t}, \mathcal{X}_{-i}, \theta\right)$, the optimal parameters $\theta=\{\gamma,\alpha_{1}, \alpha_{2}\}$, for an under consideration sequence of vehicles' states, can be obtained using the conjugate gradient optimization method as proposed in~\cite{Rasmussen:GP}.

In our settings, GPS information is converted into Earth-Centered, Earth-Fixed format (ECEF). The ECEF format is offset to the center of the vehicle on the ground and then transformed into East-North-Up (ENU) coordinates.
% In our settings, GPS latitude, longitude and elevation have been converted into the ENU coordinate system. 
As the heading and longitudinal speed of the vehicle are treated as two independent time series, we apply the GP regression technique to model and forecast the vehicles' speed and heading, and use the forecasted speed and heading to predict the position, we particularly call this approach indirect prediction. Given that the parameters of speed and heading kernel functions are obtained by optimizing the Leave-One-Out objective function using $m-$most recent observations of speed and heading data, the predictive distribution of future speed and heading values, $\mathcal{S}^{\ast}$ and $\mathcal{H}^{\ast}$, conditioned on having observed speed and heading values, $\mathcal{S}^{obs}$ and $\mathcal{H}^{obs}$, at time stamps $\boldsymbol{t}$ can be derived as
\begin{multline}\label{eq:forecastingheadingspeed}
    \Big(\mathbf{\mathcal{S}^{*}} \mid \mathbf{t^{*}}, \mathbf{t}, \mathbf{\mathcal{S}^{obs}}\Big) \sim \mathcal{N} (\mu^{\ast}_{s},\Sigma^{\ast}_{s}), \\ 
    \mspace{-75mu}\mu^{\ast}_{s}=K_{1}[\left(t^{*}, t\right)|\theta_{s}] K_{2}^{-1}[(t, t)|\theta_{s}] \mathbf{\mathcal{S}^{obs}}, \\
    \mspace{6mu}\Sigma^{\ast}_{s}=-K_{1}[\left(t^{*}, t\right)|\theta_{s}] K_{2}^{-1}[(t, t)|\theta_{s}]K_{1}^{T}[\left(t, t^{*}\right)|\theta_{s}] \\
    \mspace{18mu}+K_{3}[\left(t^{*}, t^{*}\right)|\theta_{s}],\mspace{-206mu}
    \\
    \hspace{-90pt}\Big(\mathbf{\mathcal{H}^{*}} \mid \mathbf{t^{*}}, \mathbf{t}, \mathbf{\mathcal{H}^{obs}}\Big) \sim \mathcal{N} (\mu^{\ast}_{h},\Sigma^{\ast}_{h}), \\ 
    \mspace{-75mu}\mu^{\ast}=K_{1}[\left(t^{*}, t\right)|\theta_{h}] K_{2}^{-1}[(t, t)|\theta_{h}] \mathbf{\mathcal{H}^{obs}}, \\
    \mspace{6mu}\Sigma^{\ast}=-K_{1}[\left(t^{*}, t\right)|\theta_{h}] K_{2}^{-1}[(t, t)|\theta_{h}]K_{1}^{T}[\left(t, t^{*}\right)|\theta_{h}] \\
    \mspace{18mu}+K_{3}[\left(t^{*}, t^{*}\right)|\theta_{h}],
\end{multline}
where $K_{3}$, $K_{2}$ and $K_{1}$ are the covariance matrices between future (predicted) values, observed values and observations, and future values respectively, and can be derived using \eqref{covariance func}. 
Using the predictive distributions of longitudinal speed and heading, $P(\tilde{s})$ and $P(\tilde{h})$, the position of the vehicle in the ENU coordinate system can be predicted as follows.
\begin{multline}\label{eq-positionprediction}
    \hspace{-5pt}\bar{x}\left(t_{1}\right)=x\left(t_{0}\right)+\iiint_{t_{0}}^{t_{1}} \tilde{s} \cos (\tilde{h}) \operatorname{P}(\tilde{s}) \operatorname{P}(\tilde{h}) d t d \widetilde{s} d \widetilde{h}, \\
    \bar{y}\left(t_{1}\right)=y\left(t_{0}\right)+\iiint_{t_{0}}^{t_{1}} \tilde{s} \sin (\tilde{h}) \operatorname{P}(\tilde{s}) \operatorname{P}(\tilde{h}) d t d \widetilde{s} d \widetilde{h},\hspace{7pt}
\end{multline}
Where $t_1$ and $t_0$ denote the time instance of prediction and last received BSM respectively.
In this work, since the vehicles' states were measured every $100 ms$, we calculated the predicted positions using the recursive piece-wise linear formulation.
\begin{multline}\label{eq-positionprediction2}
    \hspace{-13pt}\bar{x}\left(t_{j+1}\right)= \left(t_{j+1}-t_{j}\right)\iint \tilde{s_j} \cos (\tilde{h_j}) \operatorname{P}(\tilde{s_j}) \operatorname{P}(\tilde{h_j})  d \widetilde{s_j} d \widetilde{h_j}\hspace{-5pt} \\
    \hspace{-34pt}+\bar{x}\left(t_{j}\right)= \mathbb{E} \big[s_j\big] \mathbb{E} \big[\cos (h_j)\big] + \bar{x}\left(t_{j}\right) \\
    \hspace{-55pt}= \mu_{s_j} e^{-\sigma^2_{h_j}/2} \cos (\mu_{h_j}) + \bar{x}\left(t_{j}\right),\\
    \hspace{-22pt}\bar{y}\left(t_{j+1}\right)=\left(t_{j+1}-t_{j}\right)\iint \tilde{s_j} \sin (\tilde{h_j}) \operatorname{P}(\tilde{s_j}) \operatorname{P}(\tilde{h_j})  d \widetilde{s_j} d \widetilde{h_j}\hspace{-5pt} \\
    \hspace{-26pt}+\bar{y}\left(t_{j}\right)= \mathbb{E} \big[s_j\big] \mathbb{E} \big[\sin (h_j)\big] + \bar{y}\left(t_{j}\right) \\
    \hspace{-47pt}= \mu_{s_j} e^{-\sigma^2_{h_j}/2} \sin (\mu_{h_j}) + \bar{y}\left(t_{j}\right)
\end{multline}
where, $t_{j}-t_{j-1}=100ms$ and $t_0$ is the time instance of last received BSM. $P(\tilde{s_j})$ and $\mu_{s_j}$ are the uni-variate predictive distribution of longitudinal speed at time instance $t_{j}$ and its corresponding mean. $P(\tilde{h_j})$, $\mu_{h_j}$, and $\sigma^2_{h_j}$ are the uni-variate predictive distribution of heading at time instance $t_{j}$, its corresponding mean, and variance. These values can be calculated using equation~\eqref{eq:forecastingheadingspeed}.
The pseudo-code of our model generation scheme is illustrated in Algorithm \ref{hgp:algo}, in which $PTE$ is the Position Tracking Error and is calculated as the 2D Euclidean distance between the actual and predicted vehicle positions. Since $PTE$ sampling is dependent on the availability of actual position updates, it can be done at most at the sampling rate of GPS updates, which is 10 Hz for the SPMD dataset.
% \begin{equation}
%   PTE= \sqrt{(Erorr_X)^2 +(Erorr_Y)^2 }
% \end{equation}

% \begin{equation}
%   PTE= \sqrt{(x - \bar{x})^2 +(y - \bar{y})^2 }
% \end{equation}

%TODO update algo
\begin{algorithm}[t]
% \algsetup{indent=2em}
    \begin{algorithmic}[1]
        \caption{Training: HGP kernel bank generation} \label{hgp:algo}
        \STATE Dataset: SPMD Trips $\mathcal{T}=\{T_{1}, ...,T_{n}\}$ ; $i = 1$ \\ 
        \STATE Observe a TW history of $\mathcal{S}^{obs}$ and $\mathcal{H}^{obs}$
        \STATE Fit first model $K_{current}$ = $HGP_{fit}$ ($\mathcal{S}^{obs}$,$\mathcal{H}^{obs}$)
        \FOR{$T_i \in \mathcal{T} $}
            % \STATE $t_0\gets t_{start}$ ; $i = 0$\\
            \STATE \textbf{Read trip ${T}_i$ load training $\mathcal{S}$ ; $\mathcal{H}$}\par
            %\STATE \textbf{load $\mathcal{S}_{speed}$ ; $\mathcal{S}_{h}$; $\mathcal{S}_{ENU}$}\par
           
            \FOR{$t_{step} \in T_i $} 
            {
             % \STATE$\mathcal{S}_x=\{x_{i}, ...,x_{n}\}$ ;  $\mathcal{S}_y=\{y_{i}, ...,y_{n}\}$
          
                \STATE Use $K_{current}$
                \WHILE{$(PTE<PTE_{th})$}
                {
                
                    \begin{varwidth}[t]{\linewidth}
                    
                        % \STATE $i\gets i+1$; $t_{next}\gets t_0+i$ \par
                        \STATE $\mathcal{S}^{*}$, $\mathcal{H}^{*}$ = $HGP_{predict}(\mathcal{S}^{obs},\mathcal{H}^{obs}, K_{current}, t_{predict})$
                %      \STATE \textbf{$MP = [t_0:t_{next}]$}
                %   \STATE $[Speed, Heading]= $ 
                %           \STATE GPeval$(\textit{kernel}$, $\mathcal{S}_{speed}[MP]$, $\mathcal{S}_{heading}[MP]$)
           \STATE $(X^{*}, Y^{*})$ = $XY_{predict}(\mathcal{S}^{*},\mathcal{H}^{*})$
              \STATE $PTE$= $f_{PTE}(X^{*}, Y^{*})$
                    \end{varwidth}
                }
                \ENDWHILE 
                
                   \FOR{$K_i \in \mathcal{K} $}
                     
                    % \STATE $[Speed, Heading]= $
                        %   \STATE GPeval$(\textit{kernel}$, $\mathcal{S}_{speed}[MP]$, $\mathcal{S}_{heading}[MP]$)
                           
                        %   \STATE $(X^{predicted}_k, Y^{predicted}_k)$ =getXY($Speed,Heading$)
                            %   \STATE $PTE_k$=getPTE($X^{predicted}_k, Y^{predicted}_k$)
                        \STATE $\mathcal{S}^{*}$, $\mathcal{H}^{*}$ = $HGP_{predict}(\mathcal{S}^{obs},\mathcal{H}^{obs}, K_{current}, t_{predict})$
                          \STATE $(X^{*}, Y^{*})$ = $XY_{predict}(\mathcal{S}^{*},\mathcal{H}^{*})$   
                          \STATE $PTE_{K_i}$= $f_{PTE}(X^{*}, Y^{*})$
                        \ENDFOR
                         \STATE $PTE_{min} =\textbf{min}(PTE_{K_i})$
                        %  \STATE $\underset{K_{i} \in \mathcal{K}}{\operatorname{argmin}} PTE (K_i)$
                         \IF{$(PTE_{min}<PTE_{th})$}
                        \STATE\textbf{ Load kernel, $K_{current} = \underset{K_{i} \in \mathcal{K}}{\operatorname{argmin}} PTE (K_i)$}
                    
                         \ELSE
                         \STATE Observe a TW history of $\mathcal{S}^{obs}$ and $\mathcal{H}^{obs}$
        \STATE Fit new model $K_{current}$ = $HGP_{fit}$ ($\mathcal{S}^{obs}$,$\mathcal{H}^{obs}$)
        
                    %         \STATE \textbf{$PH= [t_{next}-TW:t_{next}]$}
                    %   \STATE ${kernel}_{new}$=GPfit($\mathcal{S}_{speed}[PH]$, $\mathcal{S}_{heading}[PH]$)\par
                         
                          \STATE \textbf{Add $K_{current}$ to $\mathcal{K}$ bank} 
                         \ENDIF
                %\STATE \textbf{create a new kernel}  

                % \par
                
                % \begin{varwidth}[t]{\linewidth}
                   
                %     % \STATE $PTE_{min} \gets \infty$;  $t_0 \gets t_{next};$ $i \gets 0$\
                % \end{varwidth}
            }
            \ENDFOR
        % \STATE $j\gets j+1$;\par
        \ENDFOR
    \end{algorithmic}
\end{algorithm}

In the algorithm, the vehicles' states (speed, $\mathcal{S}$ and heading, $\mathcal{H}$) for all $N$ available trips ($T_i$ , $i=1, ..., N$) are loaded one after another. A kernel ( $K_i \in \mathcal{K}$) is selected and used to predict the time series at each time-step, the kernel $K_i$ refers to the speed and heading kernels, the speed kernel is used to predict $\mathcal{S}^{*}$ and the heading kernel to predict $\mathcal{H}^{*}$. When a kernel is selected, its prediction accuracy is evaluated at each time step ahead. $HGP_{predict}$ is used to predict with the current kernel $K_{current}$ from the bank of kernels. As long as the kernel predicts the future positions with a $PTE$ less than the threshold ($PTE_{th}$), it remains as the selected kernel for the model and keeps using this kernel until its prediction error exceeds the threshold ($PTE >PTE_{th}$). The size of the time interval, in which the latest selected model remains in use, is called Model Persistency ($MP$). At this moment either another kernel from the kernel bank is selected for predicting the position or, if none of the available kernels could satisfy the $PTE_{th}$, a new one is created and added to the bank $\mathcal{K}$, $HGP_{fit}$ is used to learn the new model. Finally, we update $K_{current}$ and continue the prediction with the updated kernel. 

\begin{figure*}[t]
 \centering
 \includegraphics[width=.8\textwidth]{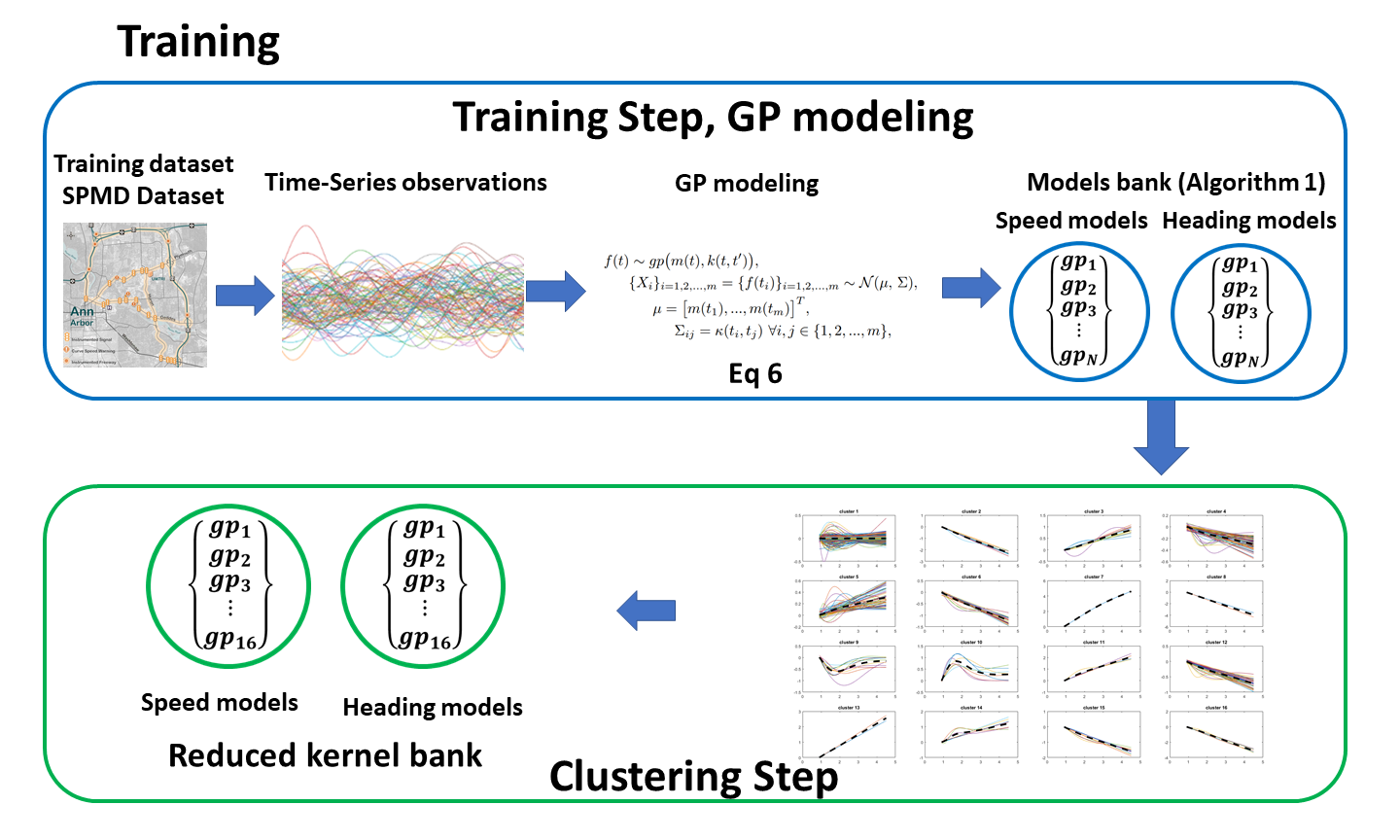}
     \caption{Offline training and clustering.}
     \label{Figure:cluster}
%  \end{figure}
 \end{figure*}

After the full kernel bank is created, the models are clustered to obtain a reduced-size kernel bank of vehicle behaviors, the limited-size kernel bank suffices to model the joint vehicle’s dynamic and behavioral pattern of the driver. During forecasting, the model with the highest likelihood is chosen from the reduced kernel bank. Figure \ref{Figure:cluster} shows a diagram of the training procedure explained before.

\subsection{Forecasting}

% \subsubsection{Using Pre-learned Kernels Bank}
Considering that the pre-learned reduced-size kernel bank is available at HV, every time the HV receives a new BSM from RV, it can choose a model from this kernel bank to forecast the trajectory (states) of the RV up until it receives the next BSM. For this purpose, assume that during the past $T$ seconds, HV received BSMs from RV at time instances $\mathbf{t}$ and the value of speed and heading of the RV at these time instances (from received BSMs) are denoted by $\mathcal{S}$ and $\mathcal{H}$. The best model from the bank, with the highest likelihood, is selected by using the following equation.

%TODO explain clustering
%  \begin{figure}[h]
 
%
%TODO check this
% \begin{equation}
% \begin{aligned}
%   \theta^{*}_{s}=\underset{\theta_{s} \in \text { speed bank }}{\operatorname{argmax}} \log p\left(\theta_{s} \mid \mathcal{S}, \mathbf{t}\right)=-\frac{1}{2} \mathcal{S}^T K_{S}^{-1}[\left(\mathbf{t}, \mathbf{t}\right)|\theta_{s}] \mathcal{S}
%   \\-\frac{1}{2} \log \left|K_{S}[\left(\mathbf{t}, \mathbf{t}\right)|\theta_{s}]\right|-\frac{n}{2} \log 2 \pi
%   \end{aligned}
% \end{equation}
%
% \begin{equation}
% \begin{aligned} 
%   \theta^{'*}=\underset{\theta^{'} \in \text { bank }}{\operatorname{argmax}} \log p\left(\theta^{'} \mid h^{*}, t^{*}\right)=-\frac{1}{2} h^{* T} K_{h}^{-1} h^{*}
%   \\-\frac{1}{2} \log \left|K_{h}\right|-\frac{n}{2} \log 2 \pi
%   \end{aligned}
% \end{equation}
%
\begin{multline}
\label{eq-,odelselection}
 \hspace{5pt}\theta^{*}_{s}=\underset{\theta_{s} \in \text { speed bank }}{\operatorname{argmax}} \log p\left(\theta_{s} \mid \mathcal{S}, \mathbf{t}\right)=-\frac{n}{2} \log 2 \pi\\
 \hspace{30pt}-\frac{1}{2} \mathcal{S}^T K_{S}^{-1}[\left(\mathbf{t}, \mathbf{t}\right) \mid \theta_{s}] \mathcal{S}-\frac{1}{2} \log  det \big(K_{S}[(\mathbf{t}, \mathbf{t}) \mid \theta_{s}]\big)   \\
  \hspace{-20pt}\theta^{*}_{h}=\underset{\theta_{h} \in \text { heading bank }}{\operatorname{argmax}} \log p\left(\theta_{h} \mid \mathcal{H}, \mathbf{t}\right)=-\frac{n}{2} \log 2 \pi \\
  \hspace{30pt}-\frac{1}{2} \mathcal{H}^T K_{H}^{-1}[\left(\mathbf{t}, \mathbf{t}\right) \mid \theta_{h}] \mathcal{H} -\frac{1}{2} \log  det \big(K_{H}[(\mathbf{t}, \mathbf{t}) \mid \theta_{h}]\big)  
\end{multline}
Where $\theta_{s}$ and $\theta_{h}$ are the hyperparameters of the speed and heading models in the bank, n is the size of vector $\mathbf{t}$ and $K_S$ and $K_h$ are the covariance matrix of speed and heading observations given hyperparameters $\theta_{s}$ and $\theta_{h}$ and can be constructed using equation~\eqref{covariance func}. $\theta^{*}_{s}$ and $\theta^{*}_{h}$ are hyperparameters of the selected speed and heading models with the highest likelihood. Using the selected model, we can predict the future values of speed and heading and consequently predict the position by utilizing \eqref{eq:forecastingheadingspeed} and \eqref{eq-positionprediction2}.

\begin{figure*}[h]
 \centering
 \includegraphics[width=.8\textwidth]{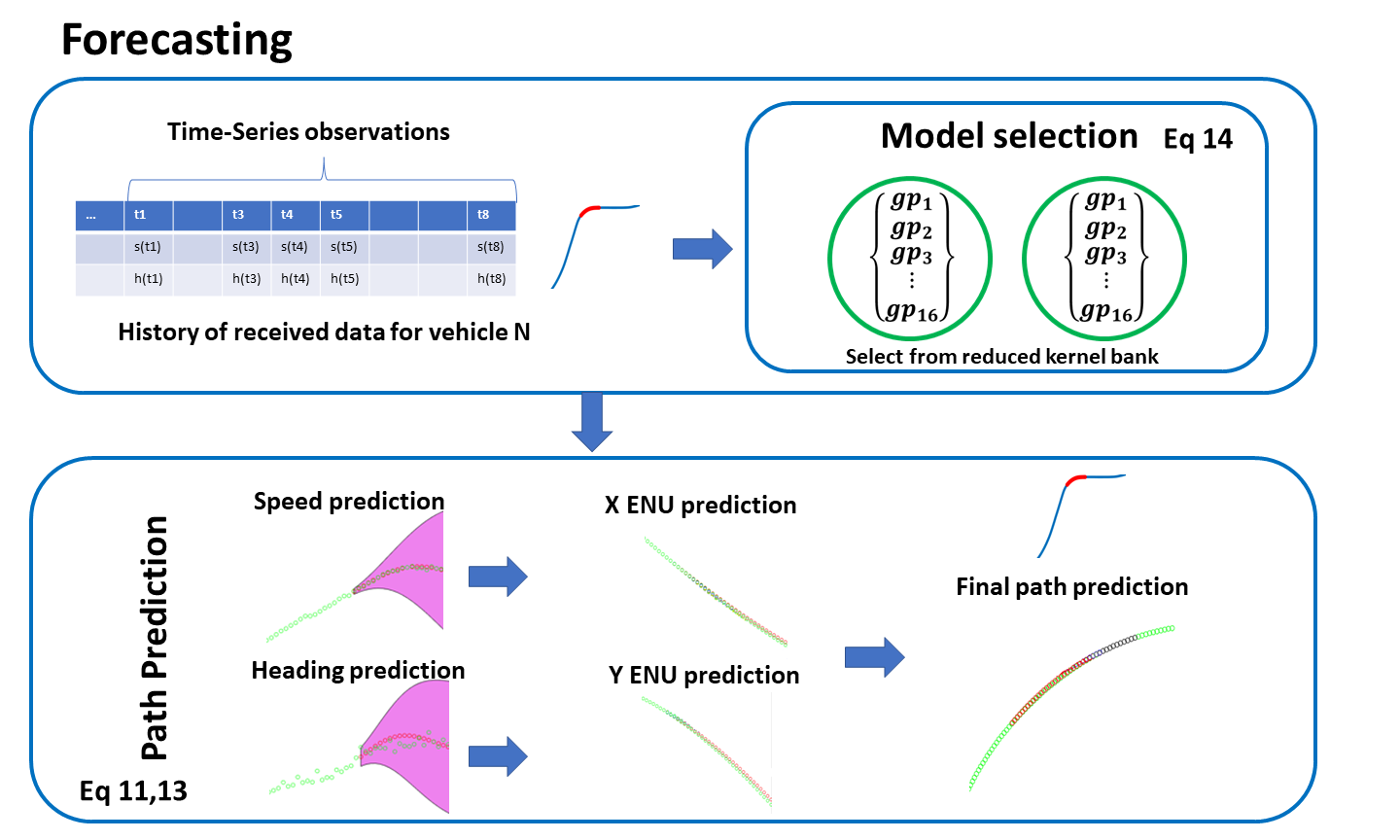}
     \caption{Path prediction example using HGP.}
     \label{Figure:forecast}
 \end{figure*}

Figure \ref{Figure:forecast} depicts how the final prediction is obtained from the history of time-series observations. After the prediction model is selected, speed and heading are predicted and used to predict X and Y and the final path.
When packets are not received, the time series history is used to select the corresponding model. If there is no model in the reduced kernel bank that meets the required $PTE_{th}$ a new model is created, added to the bank, and selected for prediction. Next, the selected model is used for forecasting the RV’s states and keeping track of the RVs even when the data is not received. 

% The Figure \ref{Figure:forecast} shows the process for the on-the-fly model selection and vehicle position forecasting. For instance, when packets are not received , the last 5 received points are use to fit a new GP model, using the pre-defined kernel and training window. Next, the model created is compare with the models in the saved model bank and if a similar model is found (based on the similarity criteria), then that model is chosen , if not the new created model is chosen and added to the bank, the chosen model is used for forecasting the RV’s states and keep track of the RV even when the data is not received. 

  \begin{figure}[h]
 \centering
 \includegraphics[width=.4\textwidth]{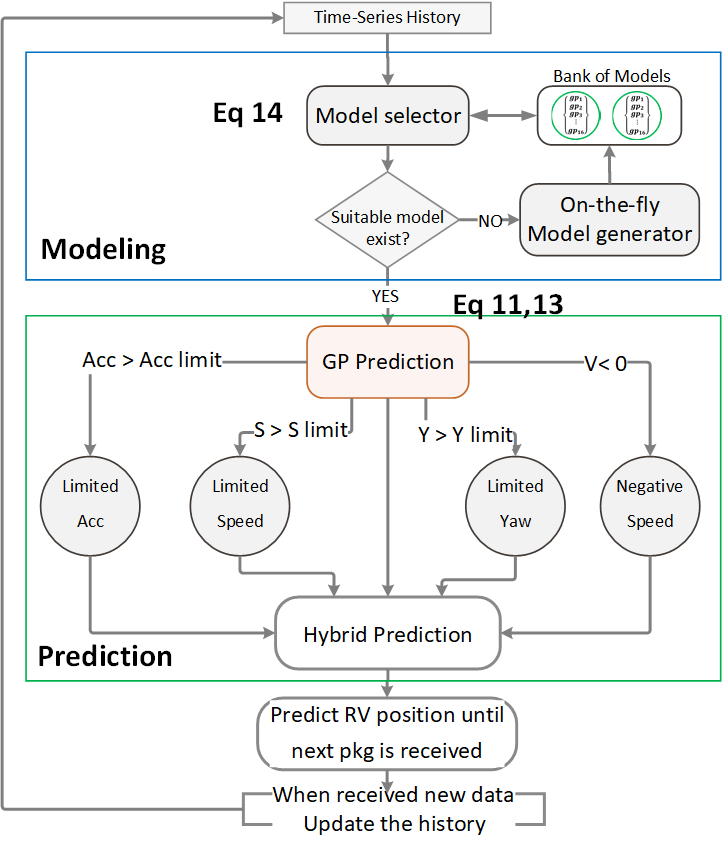}
     \caption{On-the-fly model Forecasting Diagram.}
     \label{Figure:prediction_diagram}
 \end{figure}
 
%  \begin{figure}[h]
%  \centering
%  \includegraphics[width=.48\textwidth]{Figsold/best_model.eps}
%      \caption{on-the-fly model selection inside DOM }
%      \label{Figure:best_model}
%  \end{figure}

\begin{figure*}[t]
\centering
 \includegraphics[width=1\textwidth,trim={40mm 0mm 25mm 15mm},clip]{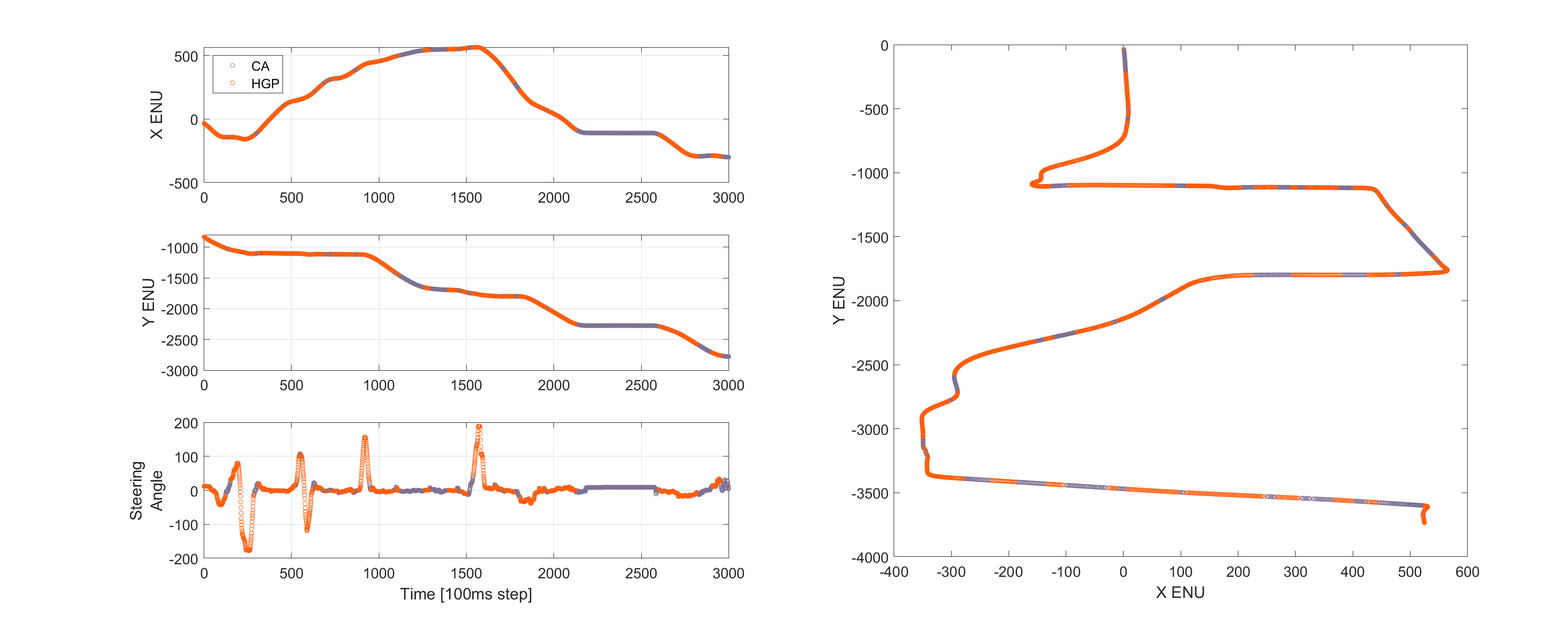}
\caption{Performance of our HGP scheme against constant acceleration (CA). The figure (using different colors) shows the moments when CA over-performs HGP (in terms of $PTE$) on the tracking accuracy while remaining below the threshold. (left) shows the performance by timesteps, (right) shows the performance in 2D ENU coordinates.}
\label{fig:hybrid_policy}
\end{figure*}

GP predictions are swapped with constant speed and constant acceleration predictions if a specific criterion is met for avoiding divergence of speed and acceleration predictions to infinity or to unfeasible values when the time interval between two consecutive BSMs is large and we must rely on prediction for a long period. Figure \ref{Figure:prediction_diagram} shows the diagram for the hybrid prediction procedure. Different motion models are used based on physical constraints in order to increase the accuracy and robustness of the prediction. This hybrid procedure probes to be advantageous, especially in situations where the assumptions of the other models are no longer valid. For visualization, the results of the hybrid approach against constant acceleration on a single trip are shown in Figure \ref{fig:hybrid_policy}. 
% For a longer than usual delay between two consecutive packets, the prediction might be highly erroneous from the CS assumption. For the CA assumption, a long delay in message reception can diverge and push the predicted speed to infinity~\cite{yfallah:mbcsyscon}. 

 \section{Experimental Setup and Simulation Results}
% Inside DOM prediction is performed when packets are not received. If a packet is received it will be processed and the information will be updated, but in absence of packet after a definite time interval, prediction will be performed (Figure \ref{Figure:estimation}). Data fusion and cooperative perception are among the capabilities of the proposed DOM architecture but are not tested here. 
We study the impact of communication uncertainties on the proposed CA-TC and compare it with the other architectures, demonstrating that the proposed prediction system outperforms the baselines.
% It is shown that our proposed prediction scheme has superior performance when compared with other approaches (baselines).

\subsection{Data Preprocessing}
% For validation, in order to make our test comprehensive we used three different datasets. First, we use SUMO simulation data to recreate a real scenario. The second dataset is the 100-car dataset \cite{Neale2005}. Finally, we use the SPMD dataset, to compare our GP-based scheme with the models that performs the best in the previous experiments.
We evaluate three separate datasets to validate the proposed architecture. To begin, we replicate realistic scenarios using SUMO simulation data. Then we use the 100-car dataset~\cite{Neale2005} and lastly, we evaluate our approach using the SPMD dataset.

\noindent
\textbf{SUMO Dataset.}
% First, we create a dataset from the SUMO simulator. The mobility logs generated are used to test our architecture, we created logs containing information such as timestamps, vehicle IDs, GPS positions, velocities, and other required data fields. Based on the transmission frequency of the vehicles (e.g. 10Hz), each vehicle log file contains a certain number of BSM packets for different timestamps. SUMO allows testing any real road scenario without the necessity of real data collection which is a costly and time-consuming task. The scenarios are chosen and exported from a map API such as OpenStreetMap and fed to the SUMO simulator, where we generate the scenarios with the desired vehicle density and behavior, and then used the designed scenarios to test our system and the CVS applications' performance.
We first evaluate our approach using the created SUMO dataset. We generate logs that contained information such as timestamps, vehicle IDs, GPS coordinates, and velocities and create the BSM packets for different timestamps based on the standard BSM transmission frequency (i.e. 10Hz). SUMO enables the testing of any real-world road scenario without the need for costly and time-consuming real-world data collecting. The designed scenarios are carefully chosen from OpenStreetMap and exported into the SUMO simulator, where we construct situations with the required vehicle density and behavior, and then utilize the intended scenarios to evaluate the performance of our system and CVS applications.

\noindent
\textbf{100-Car Dataset.}
% Second, we use the dataset from Virginia Tech Database (100-car dataset). This dataset has trajectories for over 800 real scenarios, some of them representing near-crash or actual crash scenarios. The data is used for a lead vehicle (RV), and the following vehicle (HV) movement is created using the car following the model of MITSIM~\cite{Yang2007}. The resulting vehicle movement is then fed to the system to measure the performance. As a result, we have trajectories for two vehicles (one RV and one HV), in which the RV trajectories are from real data, and HV’s are created using the car-following model.
We utilize the Virginia Tech Database dataset (100-car dataset). This dataset contains trajectories for over 800 real-world instances, some of which are near-crash or real crashes. The data is utilized to build a lead vehicle (RV) and a following vehicle (HV) movement using a car-following model as described in~\cite{Yang2007}. The subsequent vehicle mobility is used to evaluate the system's performance. As a consequence, we have trajectories for two cars (one RV and one HV), with the RV trajectories being based on real data and the HV trajectories being generated using the car-following model.

\noindent
\textbf{SPMD Dataset.}
% Finally, we use the SPMD Dataset~\cite{SPMD:Data} that is a well-known data set with a wide variety of trips collected over different urban roads with different types of vehicles and drivers. This characteristic makes SPMD a comprehensive data set that represents a diverse range of driver/vehicle behaviors and maneuvers. SPMD dataset is composed of information collected through two different settings of Data Acquisition Systems (DAS-1 and DAS-2) in Ann-Arbor, Michigan. These systems provide different in-vehicle information logged from CAN, such as longitudinal velocity and acceleration, yaw rate, steering angle, turning signal status, etc., along with the vehicle GPS information over the whole trip duration. 
Finally, we evaluate the performance of our system using the SPMD ~\cite{SPMD:Data}, which is a well-known data set that contains a wide range of trips recorded on various metropolitan roads by various sorts of cars and drivers. SPMD is a comprehensive data collection that reflects a wide variety of driver/vehicle actions and movements. The SPMD dataset provides various in-car data logged through CAN, such as longitudinal velocity and acceleration, yaw rate, steering angle, turning signal status, etc., as well as vehicle GPS data for the duration of the complete trip duration.
% In the case of the SPMD dataset, to select a diverse set of trajectories, we ranked all the trips in this dataset based on our experimental criteria and then chose the 50 richest and most diverse trips. 
To encourage the selection of a diverse set of trajectories, all the trips from the SPMD dataset are first ranked based on the experimental criteria and then 50 of the richest and most diverse trips are chosen. 
All trips are ranked on the merits of trip duration, number of successful lane changes, number of aborted lane changes, number of times a vehicle used a turning signal in a trip, number of times a vehicle stopped in a trip, and standard deviation of vehicles' yaw rate, steering angle, GPS heading and longitudinal acceleration. These 50 richest trips, which include more than $300,000$ sample points, are plotted, and visually inspected to confirm the proper selection.

The logs from the datasets (SUMO, 100Car, and SPMD) are used to recreate different communication scenarios. The logs are transferred to an OBU, replayed, and transmitted over the air by using the RVE that we have developed jointly in collaboration with industry partners for this purpose~\cite{Shah2019}. By leveraging hardware-in-the-loop, the RVE allows the joint study of CVS applications and their underlying communication system in real-time. As such, it serves the important purpose of validating the designed architecture.

\subsection{Prediction Baseline Models}
% NEW R
In order to evaluate the performance of the Hybrid GP prediction and demonstrate the capabilities of the architecture, different prediction schemes are implemented. As a baseline, we compare \textbf{BSM-dependent} (No estimation), prediction with Constant Speed (\textbf{CS}), Constant Acceleration (\textbf{CA}), Kalman Filter (\textbf{KF}), Auto Encoder (\textbf{AE}),  Long short-term memory (\textbf{LSTM}),  and the proposed hybrid GP-based scheme (\textbf{HGP}). In the next section, we describe the tested baselines. 

For AE, LSTM and GP prediction, the $(x,y)$ predictions can be obtained following two different approach, i.e, direct (\textbf{D}) or indirect (\textbf{I}). In the direct  approach, $(x,y)$ are treated as two independent time series, and the history of $(x,y)$ is used to learn  models for $(x,y)$, producing direct predictions of futures $(x,y)$. Differently, in an indirect approach, the heading and longitudinal speed of the vehicle are treated as two independent time series and the history of heading and longitudinal speed are used to learn models to forecast the vehicles' speed and heading. The predicted heading and speed are used to indirectly predict the future position $(x,y)$. In our experiments, when we refer to AE, LSTM, and HGP, the indirect approach is used. When the direct approach is used we a referred to as AE-D, LSTM-D, and HGP-D.

% NEW R
\subsubsection{Kinematic Models}
 
The BSM-dependent design (i.e. no prediction) keeps the last received position acquired from the BSM of that RV until the HV receives a new BSM from that RV. Although it is simple to implement and reduce the computation complexity of the system, it is not a realistic assumption and can lead to huge $PTE$ and the consequent drastic reduction in the performance of safety applications. The time update equation is 
$X(t+1)= X(t)$, where $X(t)$ is the position of vehicle at time t.
 
In the CS scheme, we assume that the vehicles' speed will remain unchanged during each inter-packet gap (IPG). The algorithm uses the speed from RV's last received BSM until it receives the next packet. This assumption is reasonable when the IPG is small and the HV is receiving BSMs at regular intervals. The time update equations are as below.
\begin{gather}
 X(n)
 =
 \begin{bmatrix}
   x(n) \\
   s(n)\\
   a(n) \\
   \end{bmatrix}
\end{gather}
\begin{gather}
 X(n+1)
 =
 \begin{bmatrix}
   1 & T_{s} & 0 \\
   0 & 1 & 0\\
   0 & 0 & 0 \\
   \end{bmatrix} X(n)
\end{gather}
Where $X(t)$ is the vector of vehicle states at time t and $T_{s}$ is the sampling time. This experiment assumes that $T_{s}=100ms$, which is the standard message transmission rate according to the IEEE J2735 standard. In an event of the reception of a new packet, the measurement update takes place.

The CA scheme is similar to the former approach, except that we assume acceleration to remain constant during the IPG duration. The time update mechanism follows the equation below:
\begin{gather}
 X(n+1)
 =
 \begin{bmatrix}
   1 & T_{s} & \frac{T_{s}^2}{2} \\
   0 & 1 & T_{s}\\
   0 & 0 & 1 \\
   \end{bmatrix} X(n)
\end{gather}
%A corner case of this method is a high packet loss scenario. Mathematically, the estimated speed may reach infinity for a long delay in reception, in those scenarios CA with limits and fading acceleration could help. 

\subsubsection{Kalman Filter Model}
The KF is essentially a set of mathematical equations that implement a predictor-corrector type estimator that is optimal in the sense that it minimizes the estimated error covariance when some presumed conditions are met. Rarely do the conditions necessary for optimality actually exist, and yet the filter works well for many applications. The KF addresses the general problem of trying to estimate the state of a discrete-time controlled process, it assumes that the state at time $k$ evolves from state $k-1$ according to the state update equation as, 
% that is governed by the linear stochastic difference equations
\begin{equation}
X_{k}=AX_{k-1}+Bu_{k-1}+w_{k-1}
\end{equation}
And at time $k$ a measurement(observation) is made according to,
\begin{equation} 
y_{k}=H_{k}X_{k}+v_{k}
\end{equation}
Where $X$ and $y$ represent the state vector and measurement vector; $w_{k}$ and $v_{k}$ represent the process and measurement noise respectively. The process and measurement noises are assumed to be independent white noises with normal probability distribution with covariances $Q_{k}$ and $R_{k}$ respectively. The matrix $A$ ($nxn$) is the state transition matrix which is applied to the previous state $X_{k-1}$, the matrix $B$ ($nx1$) is the control input matrix which is applied to the control vector $u_k$, and the matrix $H$ ($mxx$) is the observation model, which maps the state space into the observed space, it relates the state $X_k$ to the measurement $y_k$.

The KF computation is divided into two parts: time update and measurement update. The time update is responsible for projecting forward the current state and error covariance estimates in time to obtain a priori estimates for the next time step. The measurement update is responsible for incorporating a new measurement into a priori estimate to obtain an improved posteriori estimate. The final prediction algorithm resembles that of a predictor-corrector algorithm. The time and measurement equations are as below:

Time update equations:
\begin{equation}
X_{k} = AX_{k-1} + Bu_{k-1} +w_{k-1}
\end{equation}
\begin{equation}
P_{k} = AP_{k-1}A^T + Q
\end{equation} 

Measurement update:
\begin{equation}
K_k = P_k H^T (HP_kH^T+R)^{-1}
\end{equation}
\begin{equation}
X_k= X_k + K_k(y_k - H X_k)
\end{equation}
\begin{equation}
P_k = (I-K_kH) P_k
\end{equation}

Particularly, as a baseline, we consider a Kalman Filter with a constant acceleration dynamic, and describe the state transition equations as follows,
\begin{gather}
 X_k
 =
 \begin{bmatrix}
   x_k \\
   s_k\\
   a_k
   \end{bmatrix}
\end{gather}
\begin{gather}
\hat{X}_{k}=
 \begin{bmatrix}
   1 & T_{s} & \frac{T_{s}^2}{2}  \\
   0 & 1 & T_{s} \\
   0&0&1
   \end{bmatrix} \hat{X}_{k-1} +w_{k-1}
%   +\begin{bmatrix} \frac{T_{S}^2}{6}\\ \frac{T_{s}}{2}\\1 \end{bmatrix} w_{n}
\end{gather}
\begin{gather}
   y_k=
 \begin{bmatrix}
   1 & 1 & 1 \\
   \end{bmatrix} \hat{X}_k+v_k
\end{gather}

When a new packet is received, both measurement and time updates are running, and when a packet loss happens the prediction is done based on only the time update. Based on a statistical analysis of the dataset the variance of the measurement noise, the R matrix, is specified as, 
$$R= \begin{bmatrix}
   10 & 0&0 \\
   0 & 1&0\\
   0&0&0.5
   \end{bmatrix}$$

\subsubsection{Autoencoder Model}
% NEw REVISION

Neural network architectures such as AE use unsupervised learning or self-supervised learning. The AE architecture can vary, but in general it consists of an encoder, which reduces the input's dimension, it maps the input $\boldsymbol{x}$ to a latent feature representation $\boldsymbol{z}$ denoted by $z = f_{w_e}(x)$, and a decoder, which attempts to recreate the original input from the lower-dimensional representation, using the latent representation $\boldsymbol{z}$ to obtain a reconstruction $\boldsymbol{y}$ of the input $\boldsymbol{x}$ , denoted by $y = f_{w_d}(z)$. The goal of training these algorithms is to be able to recreate the original input with the fewest information losses possible, by minimizing the reconstruction error between $\boldsymbol{x}$ and $\boldsymbol{y}$. As a generative model, an AE can be trained for a prediction task using a similar methodology, by training the AE to reproduce the state at time $t+1$ given the state at time $t$. In our baseline, we design an encoder that consists of 3 fully connected (FC) layers with 256, 128, and 64 units respectively. The encoder takes as input the time-series observations  and outputs the internal representation that is passed to the decoder, the decoder consists of a symmetric version of the encoder. For multiple-step predictions, the predictions are used in a chain to generate subsequent predictions.
% Finally, the predictive AE is trained by minimizing the MSE

\subsubsection{LSTM Model}
An LSTM is a kind of Recurrent Neural Network. LSTMs are particularly helpful when using time series or sequential data due to their capacity to learn long-term dependencies. The LSTM network is able to learn a function that maps the time series history to feature predictions. For the LSTM baseline in this paper, we use a model similar to~\cite{altche2017lstm} with two layers of 256 LSTM cells, ReLu activation, and dropout. The model is fit using Adam and MSE loss function. It is mainly a regression problem solved by the LSTM model with the goal to predict future positions $(x, y)$ for the neighboring vehicles.

% NEw REVISION

\subsection{Controlled Variables}
% We break down the research questions of our interest into experimental hypotheses and investigate them through our experiments and ablation studies in this section.
% \textbf{Controlled Variables.}
% % 1)---Controlled/Manipulated variables  PER and Rate Packet Error Rate (PER in \%)  and transmission rate (in Hz)
% \textbf{Performance Metrics.}
% % 1)---Performance Metrics/Measures $FCW_{accuracy}$ , and $PTE$ 
% \textbf{Controlled Variables} % Independent Variables
We conducted a set of experiments to measure the performance of our architecture and investigate the existence of a reduced kernel bank size that is capable of accurately modeling driving behaviors. Then, the feasibility of the real-time implementation of the proposed system is also analyzed in terms of computation time. Finally, the overall system performance is evaluated against the previous baselines. For that purpose, we choose appropriate controlled variables, i.e, Training Window ($\mathbf{TW}$), which is defined as the number of the latest equally spaced received samples (e.g., most recent history of heading and speed time series) utilized as the training data to generate each model, Cluster Size ($\mathbf{C_{size}}$), Packet Error Rate ($\mathbf{PER}$) and Transmission Rate ($\mathbf{Rate}$).
% the independent variable are used to explore how the $TW$ size impacts the  Model Persistency ($MP$) and affect computational complexity. 
We investigate how the ${TW}$ and ${C_{size}}$ affect the model performance. Finally, In order to study the impact of communication losses, we use ${PER}$ and ${Rate}$ and measure the performance of the different prediction schemes and safety applications under different ${PER}$ and ${Rate}$. When measuring the impact of ${PER}$, the ${Rate}$ is fixed to 10Hz and when measuring the impact of ${Rate}$, the ${PER}$ is fixed to $0\%$.

\subsection{Performance Metrics}
% \textbf{Performance Metrics} % Dependent Measures
To assess the influence of the controlled variables we select suitable performance metrics, i.e, Model Persistency ($\mathbf{MP}$), computation time ($\mathbf{t_c}$), Warning Accuracy ($\mathbf{WA}$) and Position Tracking Error ($\mathbf{PTE}$). In order to gauge the impact of $TW$, we study how $MP$ and $t_c$ change with $TW$ size. We evaluate the influence of the $C_{size}$ by measuring the performance of the system in terms of $MP$ for different $C_{size}$. Finally, to measure the overall system performance, we select two performance metrics, $WA$ and $PTE$, that while correlated, provide different insights into the advantages of our approach. $PTE$ is more related to the quantitative estimation performance and $WA$ is related to the performance of the safety applications. $WA$ and $PTE$ have been used successfully as good indicators of the performance of the CVS applications~\cite{Fallah2015,Fallah2016b}.
% To measure the overall system performance, we need suitable metrics that can tell us how the safety algorithms perform in a quantitative manner. While there is no universally agreed-upon metric, Warning Accuracy (WA) and Position Tracking Error (PTE) have been used successfully in other works as a good indicator of the performance of CVS applications~\cite{Fallah2015,Fallah2016b}. 

% \noindent
\textbf{Position Tracking Error ($PTE$).}
In all the experiments, $PTE$ describes the 95th percentiles of the error in tracking the position of RV. While it measures the quantitative performance of the prediction method, it isn't the best indicator of the application performance. For this purpose, we use $WA$, which indicates how well the safety application is performing.

% \noindent
\textbf{Warning Accuracy ($WA$).}
For the purpose of validation in this paper, we focus on the FCW application, and we study the impact of communication losses on FCW accuracy($FCW_{accuracy}$). While there are many different implementations of FCW, we use the CAMP Linear FCW algorithm~\cite{Lee2005,Kiefer2003}. The used CAMP Linear FCW algorithm is briefly described in the next section. 
%As mentioned before, in this paper, we use the FCW application and implemented the CAMP Linear algorithm.
In that sense, the warnings are based on the CAMP Linear algorithm and ground truth data is the output of the CAMP Linear algorithm when PER=0 $\%$ and $Rate$=10Hz. For $FCW_{accuracy}$, we compare the ground truth and FCW application output for different values of $PER$ and $Rates$ while different prediction schemes are used. More precisely, $FCW_{accuracy}$ is the ratio of true negatives plus true positives across all execution instances and is defined as follows:
\begin{equation}
FCW_{accuracy} = \frac{T_p +T_n}{F_p +F_n +T_p +T_n}
\end{equation}
where $T_p$ (True positive) and $T_n$ (True negative) are the numbers of correctly predicted hazards and safe indications; $F_n$ (False negative) is the number of incorrectly predicted safe indications (misidentified actual hazards), and $F_p$ (False positive) is the number of incorrectly predicted hazard indications (misidentified actual safe situations).

We compare the $PTE$ and $FCW_{accuracy}$ (Performance metrics) of the CA-TC-based architecture using different prediction schemes, while $PER$ and $Rate$ (Controlled variables)  change from 0 to 95 $\% $ and from 10Hz to 1Hz respectively. %Considering the same FCW algorithm for all tests, the $FCW_{accuracy}$ depends on the quality of the predictions. In these terms, the BSM-based method is equivalent to having no prediction since it assumes the last BSM info like the current info until the HV receives the next BSM from that RV. 

%  Different safety applications will be invoked based on a periodic TC, which process the data and provide it to the applications. In our proposal, TC data is based on the DOM information rather than individual BSMs. This characteristic enables the advantage of using prediction module while the data has been lost (due to network congestion) or it is extensively noisy. As it will be shown in the following section, this will increase the performance of CVS systems. Separation of application layer from information layer also facilitate to add new applications without the need of any major change in the architecture. A wide variety of safety application have been developed in the past few years and can be deployed on our proposed architecture. All the common safety applications such as: FCW, Electronic Emergency Brake Light (EEBL), Intersection Move Assist (IMA), Do Not Pass Warning (DNPW), Blind Spot Warning/Lane Change Warning (BSW/LCW) have been tested with the DOM-based architecture. In this paper, we analyze how FCW performs based on our architecture.
 
% In this paper, in order to take the effect of communication loss into account and assess the performance of our proposed system under more realistic conditions, 
Although our architecture is agnostic to the V2X communication technology, in our setup, we used the V2X device (DSRC-based) DENSO WSU-5900A and the RVE to evaluate the performance of the system~\cite{Shah2019}. We show the results of the analyses for the SUMO, 100-Car, and SPMD datasets by using the mobility logs together with the RVE. The RVE can imitate network congestion scenarios with different $PER$ or $Rates$.

% Finally, we compared the performance of our metrics over all the three-studied datasets, i.e, SUMO, 100-Car, SPMD. As discussed the communication analyses are studied by using the mobility logs together with the RVE using a DENSO WSU-5900A device~\cite{Shah2019}. 

% \subsection{Hypotheses} % connect variables to measures
% 1)---Connect manipulated variables and performance measures
% provide experimental results that answer the hypotheses 

\subsection{Cooperative Forward Collision Warning (FCW)}

% In this work, we use the FCW application as a test case for our experiments and implemented the CAMP Linear algorithm~\cite{Lee2005,Kiefer2003}. 
In this section, we briefly present the FCW CAMP Linear algorithm implementation used in the experiments~\cite{Lee2005,Kiefer2003}. FCW algorithms are intended to notify the driver when the present movement pattern of cars indicates that a collision is imminent. The warning should be timed such that it takes into account the driver's reaction time, but not so quickly that it causes false alarms. FCW is run repeatedly (e.g., 100 ms intervals) and at each interval is evaluated if a threat exists. The FCW CAMP Linear algorithm evaluates a “warning range” ($r_w$) using RV and HV data.
% FCW is usually executed periodically (e.g., 100 ms intervals) and at each execution, the time instance determines whether a threat exists. The CAMP Linear algorithm uses HV and RV information to calculate a “warning range” ($r_w$). This algorithm assumes that information about the position, speed, and acceleration of RV is available at the HV. 
A warning is issued if the distance between the HV and the RV is less than the $r_w$~\cite{Lee2005,Kiefer2003,Fallah2015}.

%The $r_w$ and distance between vehicles are dynamic and are continuously monitored. In the FCW algorithm, the alert timing is set up such that a warning is delivered just when it becomes necessary for the driver. The alert time is referred to as a "warning range," which varies as the speed, acceleration, and distance change.
%A collision warning should be provided to the Following Vehicle (FV) if the distance between a Leading Vehicle (LV) and the Following Vehicle (FV) gets smaller than the warning range.
%  FCW algorithms are intended to notify the driver when the present movement pattern of cars indicates that a collision is imminent. The warning should be timed such that it takes into account the driver's reaction time, but not so quickly that it causes false alarms. In our experiments we the CAMPLinear algorithm from [7] to test the performance of FCW under different frameworks, measured by $FCW_{accuracy}$. In the FCW algorithm the alert timing is set up such that a warning is delivered just when it becomes necessary for the driver. The alert time is referred to as a "warning range," which varies as the speed, acceleration, and distance change. A collision warning should be provided to the Following Vehicle (FV) if the distance between a Leading Vehicle (LV) and a Following Vehicle (FV) gets smaller than the warning range [7] [8] [10]. The FCW equations are summarized as follows, 

% \noindent
\textbf{CAMP Linear FCW algorithm.}
% The details of the algorithm are described in \cite{Lee2005,Kiefer2003,Fallah2015} and summarized as follow, 
In the CAMP Linear FCW algorithm, the $r_w$ is computed as:
\begin{equation} \label{equ:FCW}
\begin{aligned}
r_w = \mathrm{BOR} + (\mathrm{v}_{\mathrm{HV}} - \mathrm{v}_{\mathrm{RV}})t_d  + \frac{1}{2}*(\mathrm{a}_{\mathrm{HV}} - \mathrm{a}_{\mathrm{RV}})t_d^2
\end{aligned}
\end{equation}
Where $r_w$ is the warning range, $BOR$ is the Brake Onset Range; $v_{HV},v_{RV},a_{HV},a_{RV}$ are the speed and acceleration of HV and RV and $t_d$ is the driver and brake system reaction delay. The $BOR$ is computed for three different scenarios, for stationary 
or moving RV, as follows.
\begin{itemize}
  \item \textbf{Case 1:} RV stationary. 
  \item \textbf{Case 2:} RV moving at the beginning and end of the scenario.
  \item \textbf{Case 3:} RV moving at the beginning but stopping at the end of the scenario.
\end{itemize}
\begin{equation} \label{equ:FCW}
\begin{aligned}
& \textbf{Case 1: } \mathrm{BOR} = \frac{\mathrm{v}_{\mathrm{HVp}}^2}{-2*a_{req}} \\
& \textbf{Case 2: }\mathrm{BOR} = \frac{(\mathrm{v}_{\mathrm{HVp}} - \mathrm{v}_{\mathrm{RVp}})^2}{-2*(a_{req} -a_{RV})}  \\
& \textbf{Case 3: }\mathrm{BOR} = \frac{\mathrm{v}_{\mathrm{HVp}}^2}{-2*a_{req}} - \frac{\mathrm{v}_{\mathrm{RVp}}^2}{-2*a_{RV}}
\end{aligned}
\end{equation}
in witch, $v_{HVp}$  and $v_{RVp}$ are the predicted velocity of HV and RV, i.e, $v_{HVp} = v_{HV} + a_{HV}t_d$ and $v_{RVp} = v_{RV} + a_{RV}t_d$. $a_{RV}$ is the acceleration (deceleration) of RV and $a_{req}$ is the deceleration that is required at the HV for avoiding a crash and is modeled in~\cite{Kiefer2003} using real data.
% FCW in our experiments will fully rely on the information provided by the V2V network in the BSM-dependent scheme and will be prediction based in the other cases accordingly. 

\subsection{Hypotheses}
% We break down the research questions of our interest into experimental hypotheses and investigate them through our experiments and ablation studies in this section.
Based on the defined \textbf{controlled variables} and \textbf{performance metrics} we examine the following hypotheses:
\begin{itemize}
    \item \textbf{H1.} \emph{An suitable {$\mathbf{TW}$} needs to be selected in order to allow an acceptable model persistency ($\mathbf{MP}$) and meet the real-time requirements of the CVS applications measured by computation time ({$\mathbf{t_c}$})}.
    \item \textbf{H2.} \emph{We anticipate the existence of a limited size of kernel banks ({$\mathbf{C_{size}}$}) which is capable of predicting driving behavior while keeping an acceptable $\mathbf{MP}$ horizon. The kernel bank can be learned offline and still be useful during on-the-fly forecasting}.
    % TODO Connect variable and measure better
    \item \textbf{H3.} \emph{The higher the {$\mathbf{PER}$} or the lower the {$\mathbf{Rate}$} of communication, the greater the impact and benefit of using our CA-TC system, measured by {$\mathbf{PTE}$} and  {$\mathbf{FCW_{accuracy}}$}. Thus, we expect a higher $\mathbf{FCW_{accuracy}}$ when using our HGP prediction scheme when compared to the baselines}.
%?? are manipulated variables and performance measures connected clearly ???
\end{itemize}

The hypotheses are investigated through the experiments in the following sections. 

\subsection{GP Prediction Model Hyper Parameter Tuning and Implementation Details}

As described, in the Hybrid GP modeling scheme instead of working directly on the X-ENU and Y-ENU time series, the heading, and speed of the vehicle are treated as two independent time series. The GP models, which are learned from speed and heading histories are used to forecast their future value, and then the predicted values of these two variables are used to predict the position in ENU coordinate system. 
%A compound GP kernel type, composed of a linear and an RBF kernel, is selected based on our comprehensive observations.
In this section, we present the study of $TW$ and $C_{size}$ and obtain suitable values for those variables to be used for the HGP prediction system. 
% We conducted numerous experiments and investigated four different $PTE$ thresholds ($PTE_{th}$), i.e. 20 cm, 30 cm, 40 cm, and 50 cm. These values cover the range between minimum and maximum thresholds specified by the SAE J2945/1 standard. CVS applications could have different requirements for $PTE_{th}$ and in our model generation, the $PTE_{th}$ determines the moments when the current model is not valid anymore and a new model should be selected from the available kernel bank or should be generated if an appropriate model for this data does not exist in the available kernel bank by this moment. 

\subsubsection{Computational Complexity}
% According to the details of our algorithm, the complexity depends on the new model generation ratio, the clustered kernel bank size and the training window size.
To study the hypothesis \textbf{H1} we investigate the effect $TW$ in computation time and experimentally compute the computation time of our algorithm for $TW$ size from 1 to 1000. The HGP training (new kernel generation) and forecasting complexity is a function of the training window. In our framework, this factor appears whenever we need to generate a new kernel while we are running the system in real-time. The order of this complexity is heavily dependent on the implementation strategy of GP kernel training algorithms. In the simulations, we have used the "Gaussian Process for Machine Learning" MATLAB toolbox, provided by Cambridge University.
% In order to have an empirical estimation of the kernel training time complexity,
%The implementation was a single-thread and only around 4 \% of CPU power was under-utilization. 
% The approach can benefit from parallel computing during the bank kennel creation and during forecasting, as the model predictions can be computed independently in parallel. The simulations have been executed on a PC with an Intel Core i7 6700 3.4 GHz CPU with 4 cores.

\begin{figure}[t]
\centering
 \includegraphics[width=.48\textwidth]{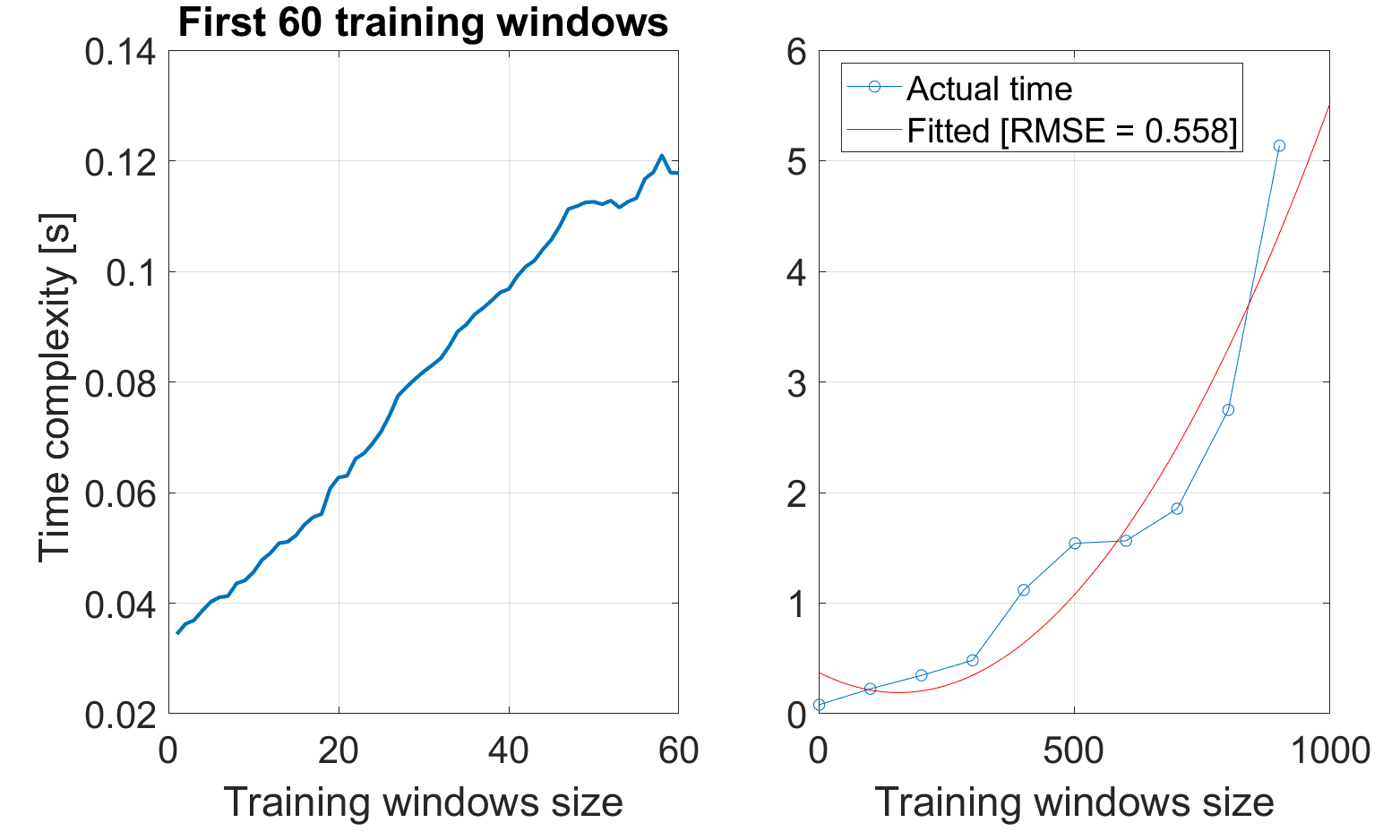}
\caption{Empirical time complexity of GP kernel training as a function of training window size.}
\label{fig:complexity}
\end{figure}

Figure~\ref{fig:complexity} shows the result of the experimental estimation, in the right plot the blue curve shows the actual spent time and the red one is the quadratic approximations, as shown for a large number of training windows, the complexity of our algorithm is quadratic ($\mathcal{O}(n^2)$). The Root Mean Square Error (RMSE) values for the ﬁtted curve are also presented in the ﬁgure.
In the left plot, a plot of the computation time for training windows size up to 60 is presented. As shown we can train up to $TW$ size 40 and still be under the 100ms threshold required for safety applications. The simulations have been executed on a PC with an Intel Core i7 6700 3.4 GHz CPU with 4 cores and 32 GB of RAM. The results in Figure~\ref{fig:complexity} validate the \textbf{H1} and allows us to select a suitable $TW$ that meets the real-time requirements of the CVS applications. Additionally, we want to mention that our approach can benefit from parallel computing during the bank kennel creation and during forecasting, as the model predictions can be computed independently in parallel, therefore the use of GPU or a CPU with additional cores will further reduce the computation time. Based on our experiments and previous work~\cite{Mahjoub}, we choose a $TW$ size of 30 time-steps (3s) and $PTE_{th} = 50cm$ for the training, model generation, and forecasting. The selected training parameters consider an acceptable path history while also meeting the computational requirement of the safety applications. 
%Figure \ref{fig:complexity} shows that the inference time of our model with a training window of size 30 still is under the 100ms requirement for the safety applications.
% PTE based on application = 50cm , Training window = 30, 3s. based on application and computaional complexity 

\subsubsection{Model Generation}
%  During model generation the vehicle's states for all 
% $N$ available trips ($T_j$ , $j=1, ..., N$) are loaded one after another. At the beginning of each trip, $t_0$ (prediction start time) is initialized and the algorithm tries to predict the time series belong to this trip using the kernels which have been created so far. When a kernel is selected, its prediction accuracy is evaluated at each time-step ahead. As long as the kernel predicts the future positions with a $PTE$ less than the threshold ($PTE_{th}$), it remains as the selected kernel for our model and keeps this title until its prediction error exceeds the threshold ($PTE >PTE_{th}$). The size of the time interval ,in which the latest selected model remains in use, is called Model Persistency ($MP$). At this moment either another kernel from our kernel bank is selected for predicting the position or, if none of the available kernels could satisfy the $PTE_{th}$, a new one is created and added to the bank. Finally, we update $t_0$ to start the prediction with the updated kernel. 

\begin{figure}[t]
\centering
 \includegraphics[width=.35\textwidth]{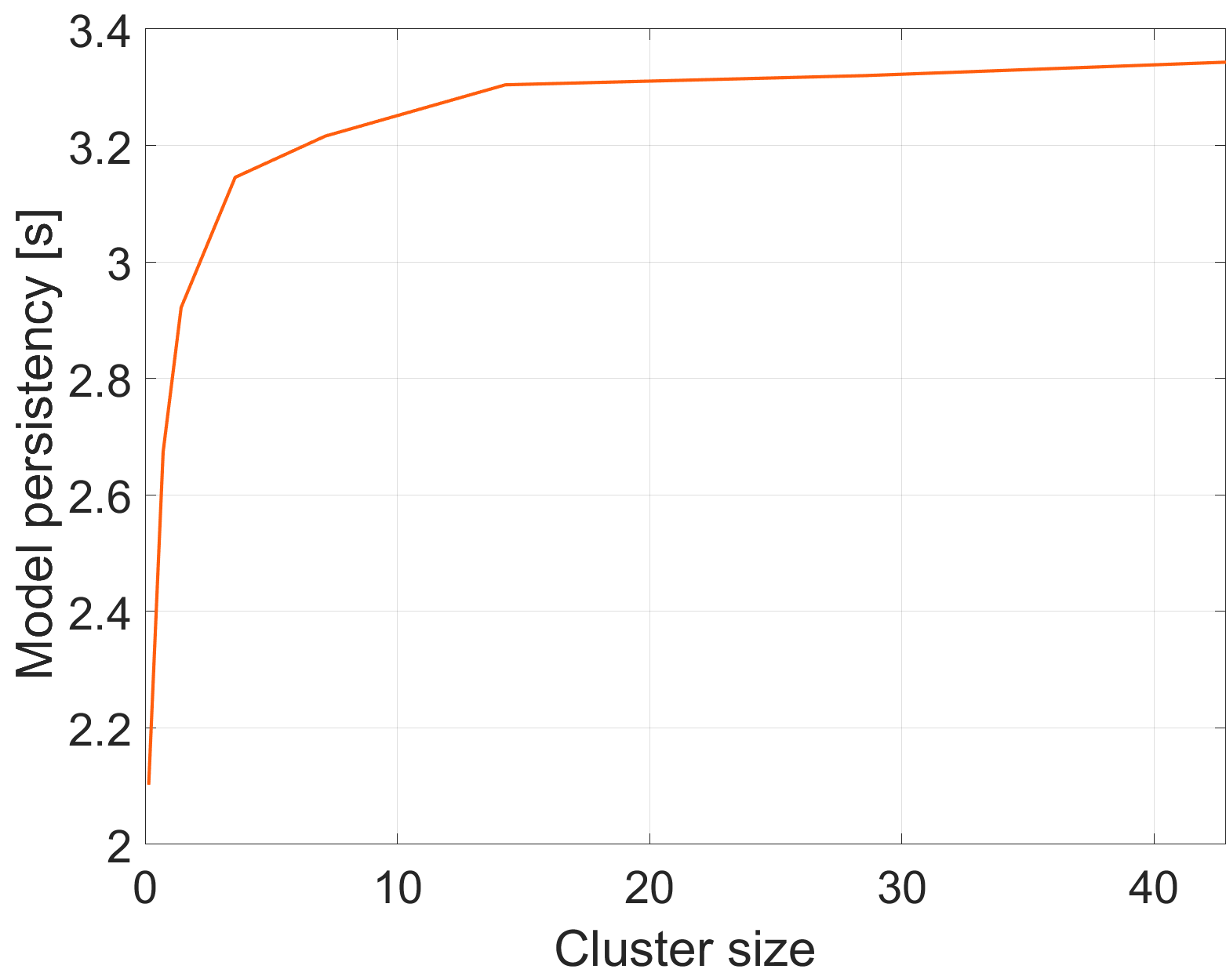}
\caption{Model persistency for different cluster sizes.}
\label{fig:cluster_size}
\end{figure}

%TODO explain the cluster algo procedure
To examine the hypothesis \textbf{H2}, we obtain the kernel bank and experiment with different cluster sizes. Following the HGP training and model generation schemes in Algorithm~\ref{hgp:algo}, we obtained 350 models, and reduce the kernel bank by clustering the kernels. For each $C_{size}$ we compute the model persistency ($MP$) when predicting future trajectories while keeping the error within the defined $PTE_{th}$. Figure~\ref{fig:cluster_size} shows how $MP$ changes for different $C_{size}$. The results support the hypothesis \textbf{H2} and confirm the existence of a reduced-size kernel bank capable of predicting driving behavior patterns. From those results, we choose a cluster size of 16 that has a reduced size while keeping an adequate $MP$, balancing complexity and performance. Therefore the models from the original kernel bank are clustered in the 16 most distinctive models to form the reduced kernel bank.

%Figure \ref{Figure:best_model} the diagram of the Forecasting phase. Forecasting is done on-the-fly. During forecasting, we first use the previous history and create a new model when data is not received
During on-the-fly mode, the most recent information is used to learn the parameters of the Gaussian process’s Kernel. 
% The time duration of the path history used to learn the models in on-the-fly mode is not fixed and it depends on the packet dropout rate and channel congestion. 
In the experiments, we use BSMs received during the last $TW$ to learn the new model.
%Therefore, the size of path history is 30, considering that the rate of transmission is 10 Hz. 
This model will be used to forecast the state of the vehicle until new information is received or the forecasted speed or acceleration of the vehicle triggers the model switching based on the algorithm described in Figure~\ref{Figure:prediction_diagram}.

\begin{figure}[t]
\centering
 \includegraphics[width=.35\textwidth]{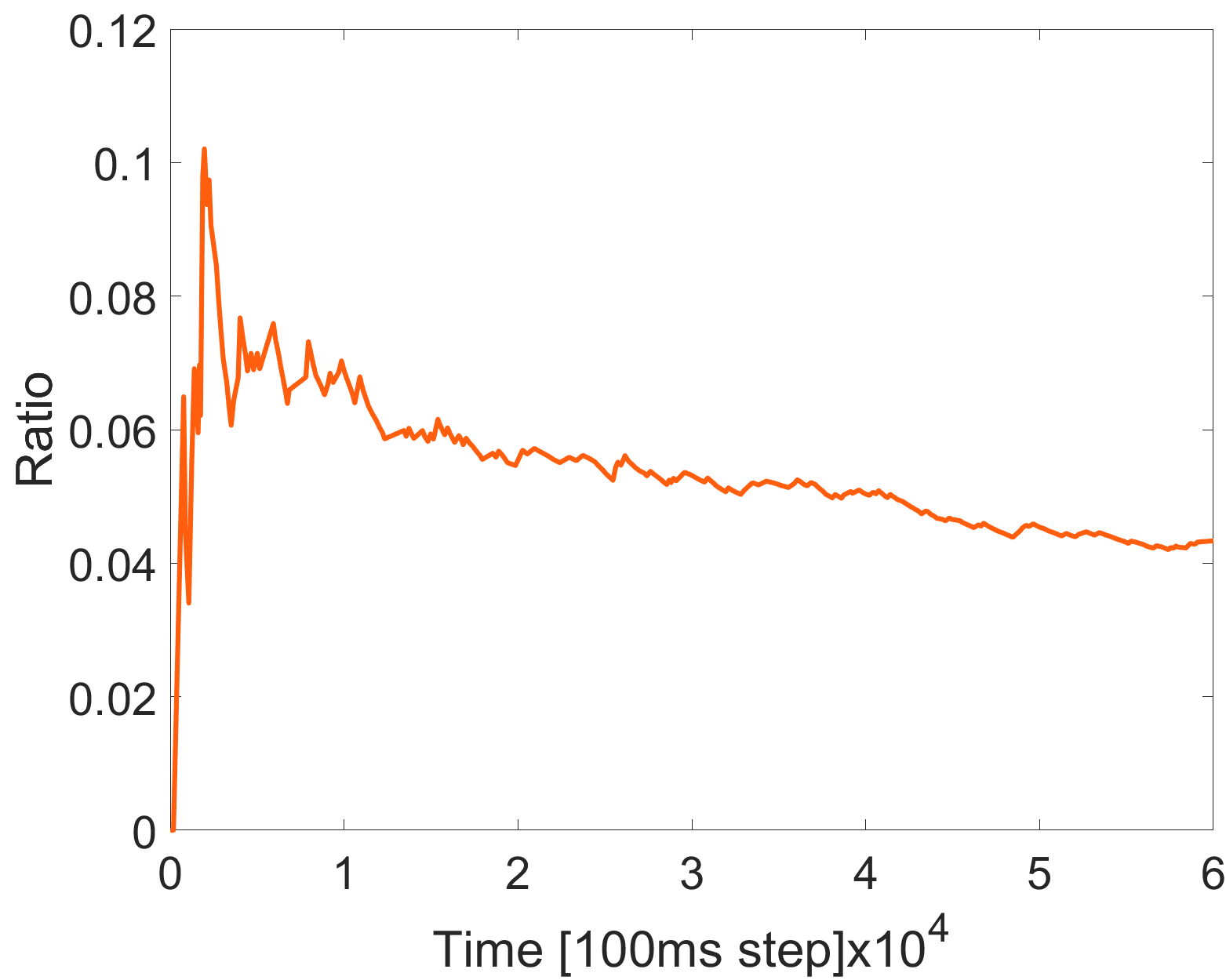}
\caption{Assessment of trained kernel bank performance in terms of required new kernel generation rate for an unforeseen set of trajectories.}
\label{fig:Assessment}
\end{figure}

Finally, to test the capability of the reduced kernel bank, we apply the trained kernel bank to a new set of 50 randomly selected trajectories. Therefore, they could be considered as a set of generic driving data. The ratio of generating new models, which has been represented in Figure~\ref{fig:Assessment}, has an increasing trend at the beginning of test time and raises to around 10\% at its maximum level, and then it converges around 4\%. This observation shows that the learned kernel bank is meaningful during the testing phase (on-the-fly forecasting), even for unforeseen data, further verifying the hypothesis \textbf{H2}.

\subsection{Overall system performance}
Finally, to investigate the \textbf{H3}, we use the parameters obtained for our HGP prediction system, i.e. $TW=3s$, $PTE_{th}=50cm$, and $C_{size}=16$, and study the impact of communication uncertainties on the performance of a selected safety application with proposed CA-TC and compare it with the baselines across multiple datasets.

\begin{figure}[h!]
 \centering
 \includegraphics[width=.48\textwidth]{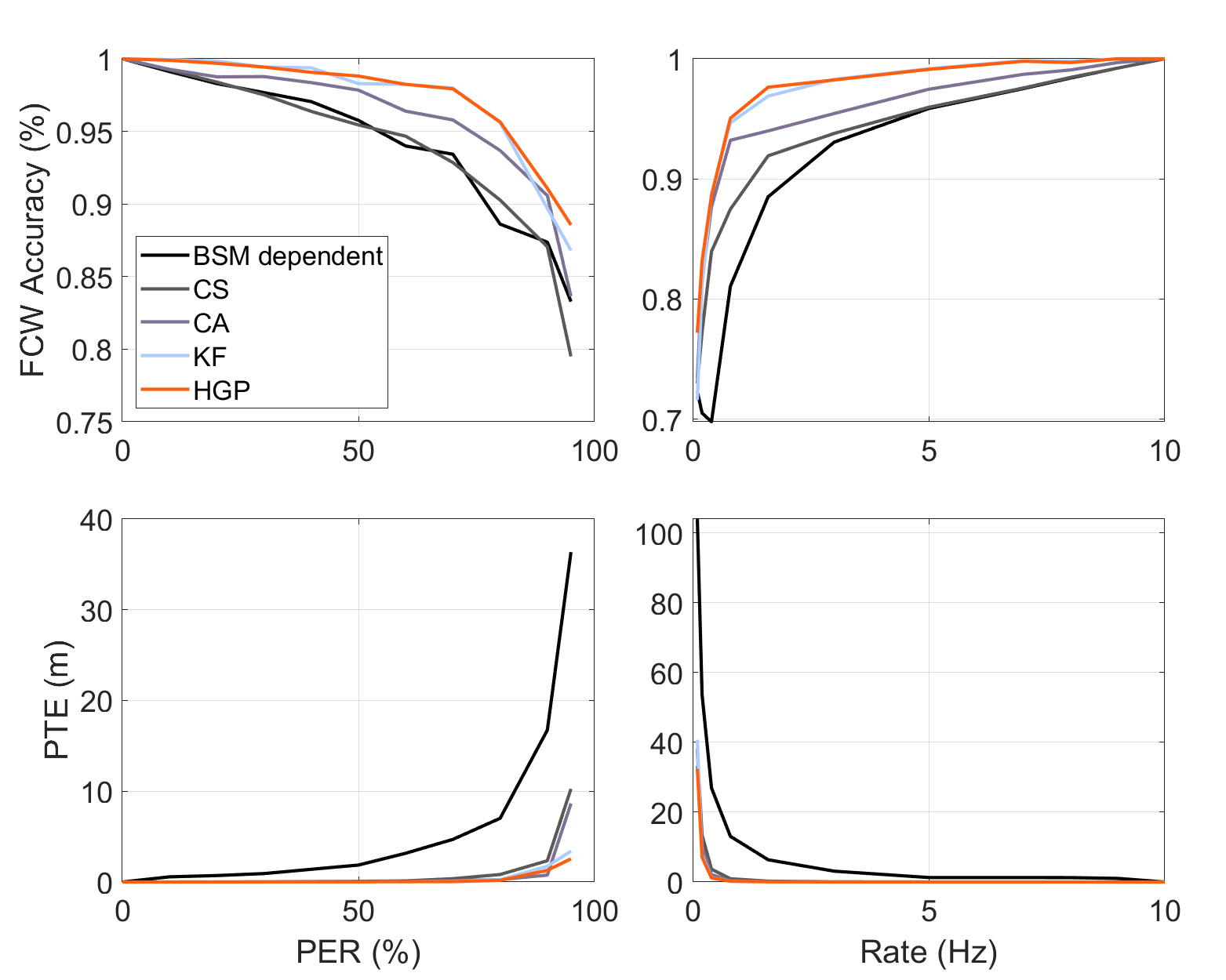}
     \caption{$PTE$ and $FCW_{accuracy}$ vs $PER$ and $Rate$ for the SUMO dataset.}
     \label{Figure:Sumo}
 \end{figure}

\noindent \textbf{Experiments with SUMO simulated data}
We first use the data generated from the SUMO simulator and compare the baseline models against our approach. Figure \ref{Figure:Sumo} demonstrates $FCW_{accuracy}$ and $PTE$ for different values of $PER$ and $Rates$ using the SUMO data. As expected, the $FCW_{accuracy}$ of the BSM-dependent baseline is the lowest, and the $PTE$ is the highest. For the SUMO simulated data, the KF model and GP perform similarly which could be a consequence of the simplistic car-following model utilized in SUMO. 
% An example of the SUMO scenario for an intersection at University of Central Florida is shown in Figure \ref{Figure:Sumo}.

%  \begin{figure}[h]
%  \centering
%  \includegraphics[width=.48\textwidth]{Figsold/sumo.png}
%      \caption{Example of a SUMO simulator scenario}
%      \label{Figure:Sumo}
%  \end{figure}

\begin{figure}[h]
 \centering
 \includegraphics[width=.48\textwidth]{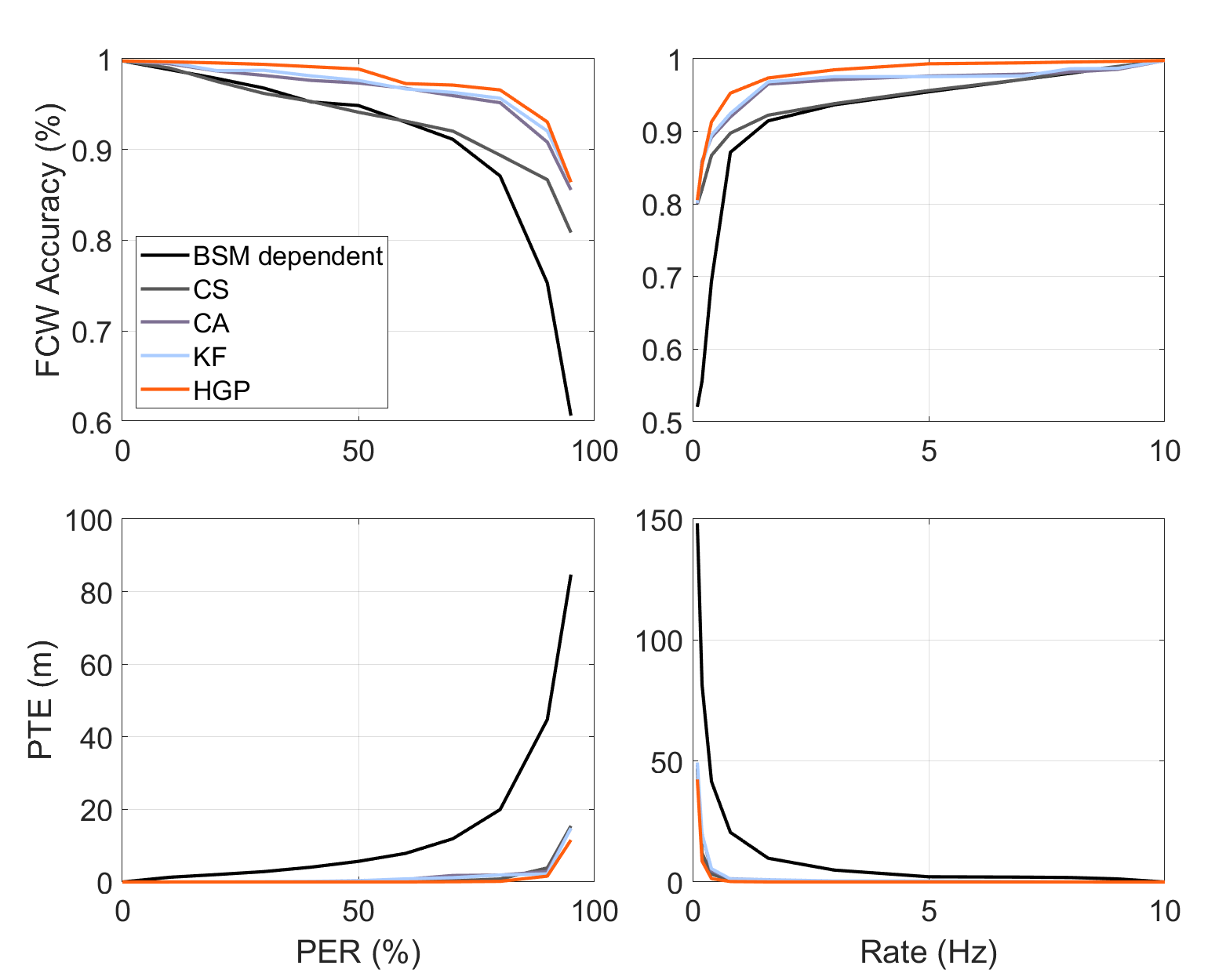}
     \caption{PTE and $FCW_{accuracy}$ vs $PER$ and $Rate$ for the 100-Car dataset}
     \label{Figure:100Car1}
 \end{figure}

\noindent \textbf{Experiments with the 100-Car Dataset}
We also evaluated the performances using the 100-Car dataset. Figure \ref{Figure:100Car1} shows how the $PTE$ and $FCW_{accuracy}$ change for the different $PER$ and $Rate$ values. Similarly, the BSM-dependent model shows the highest $PTE$ and lowest $FCW_{accuracy}$. HGP has a similar performance to KF or CA models. We speculate that the reason for having similar performances is the car-following model used to create the HV states in the 100-Car dataset. 

% Figure \ref{Figure:100Car3},\ref{Figure:100Car4} shows how accuracy reduces as PER increases; when PER is 0, accuracy is 1 (i.e. all the packets are received, and the warnings matches exactly with the ground truth).  As expected, the BSM-Based model shows the highest prediction error and lowest $FCW_{accuracy}$. 

%  \begin{figure}[h]
%  \centering
%  \includegraphics[width=.48\textwidth]{Figsold/100Car_p1.eps}
%      \caption{PTE vs PER for the 100-Car dataset}
%      \label{Figure:100Car1}
%  \end{figure}

%  \begin{figure}[h]
%  \centering
%  \includegraphics[width=.48\textwidth]{Figsold/100Car_p2.eps}
%      \caption{PTE vs Rate for the 100-Car dataset}
%      \label{Figure:100Car2}
%  \end{figure}

%   \begin{figure}[h!]
%  \centering
%  \includegraphics[width=.48\textwidth]{Figsold/100Car_p3.eps}
%      \caption{Accuracy vs PER for the 100-Car dataset}
%      \label{Figure:100Car3}
%  \end{figure}

%   \begin{figure}[h!]
%  \centering
%  \includegraphics[width=.48\textwidth]{Figsold/100Car_p4.eps}
%      \caption{Accuracy vs Rate for the 100-Car dataset}
%      \label{Figure:100Car4}
%  \end{figure}

% \subsection{Comparison of prediction methods using SPMD dataset}

\begin{figure}[h!]
\centering
\includegraphics[width=.48\textwidth]{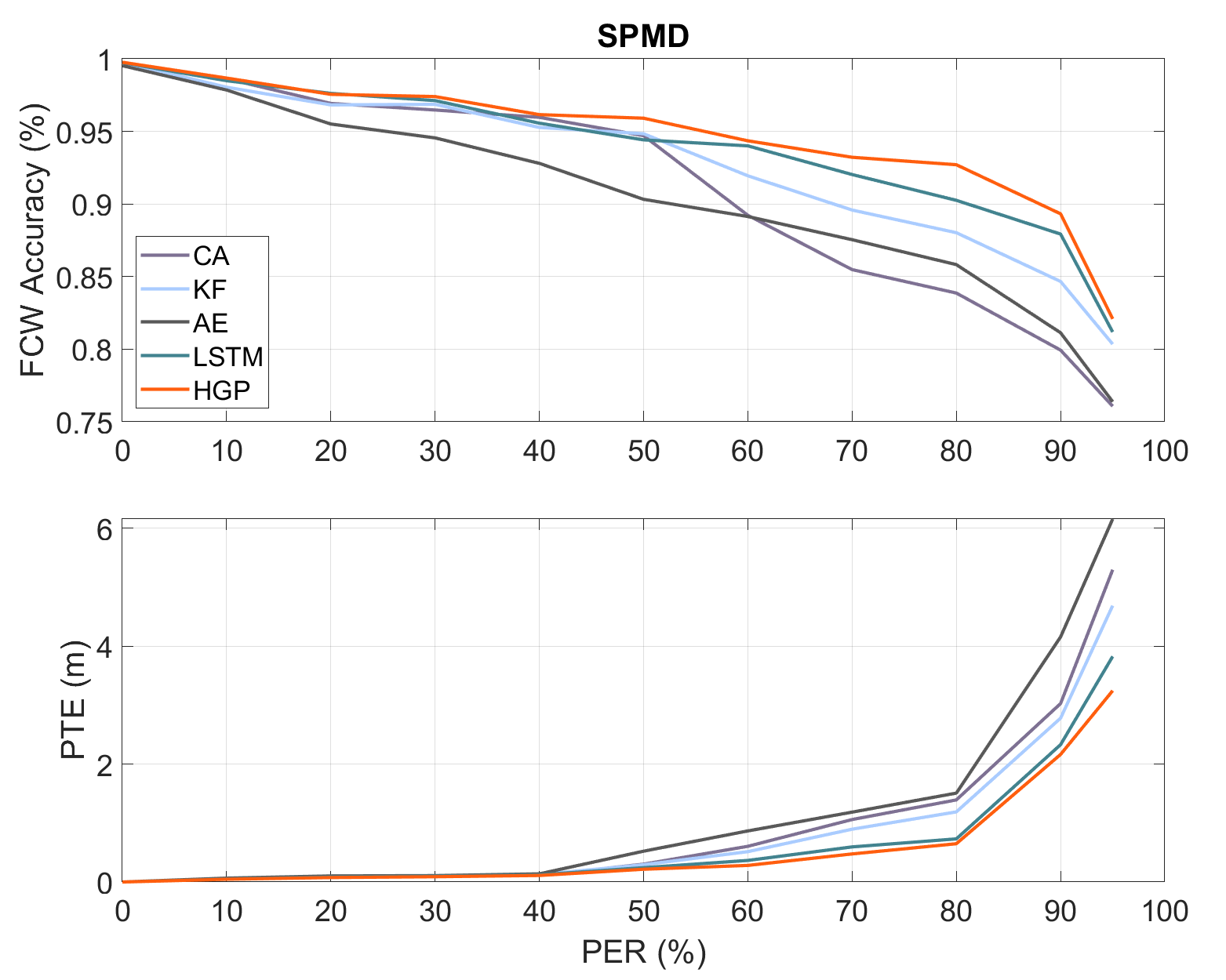}
 \caption{Comparison of CA, KF, AE, LSTM and our HGP approach performance in the SPMD dataset, measured by $FCW_{accuracy}$ and $PTE$}
 \label{Figure:All}
\end{figure}

% NEW REVIEW
\noindent \textbf{Experiments with the SPMD dataset}
Finally, using SPMD dataset~\cite{SPMD:Data}, we compare the proposed HGP method with the baselines CA, KF, AE, and LSTM. We want to mention that the AE and LSTM models require a larger amount of data and are not suitable for the SUMO and 100-Car datasets.
%This well-known data set has a wide variety of trips collected over different urban roads with different types of vehicles and drivers. This characteristic makes SPMD a comprehensive data set which represents a diverse range of driver/vehicle behaviors and maneuvers. SPMD data set is composed of information collected through two different settings of Data Acquisition Systems (DAS-1 and DAS-2) in Ann-Arbor, Michigan. These systems provide different in-vehicle information logged from CAN, such as longitudinal velocity and acceleration, yaw rate, steering angle, turning signal status, etc., along with the vehicle GPS information over the whole trip duration.
Figure \ref{Figure:All} shows the performance of all the models in the SPMD dataset. As expected, the KF model performed similarly to the CA model due to the chosen kinematic equation to model the vehicle motion in the KF method. However, the slight over-performance of KF is due to its ability to reduce measurement noise, in particular, the AE model shows poor performance for vehicle trajectory prediction. As $PER$ increases over $60\%$, the HGP approach shows its superior prediction, as it can forecast for longer prediction horizons, and capture both the dynamics of the vehicle and the driver's behavior. The difference in position tracking is notable as $PER$ reaches 90$\%$. The CA and KF approaches had $60\%$ larger $PTE$ compared to HGP. The LSTM model has a similar performance to our HPG model however LSTM models are data-hungry models and are not suitable for online on-the-fly training while new data is available, making it difficult to extract new knowledge from new situations, differently from the HGP approach that can learn new models on the fly, with a limited amount of data. The results demonstrate the superiority of our HGP approach and the benefits as $PER$ increases, confirming the hypothesis \textbf{H3}.
% NEW REVIEW

\begin{figure*}[h]
\centering
 \includegraphics[width=1\textwidth,trim={45mm 5mm 45mm 5mm},clip]{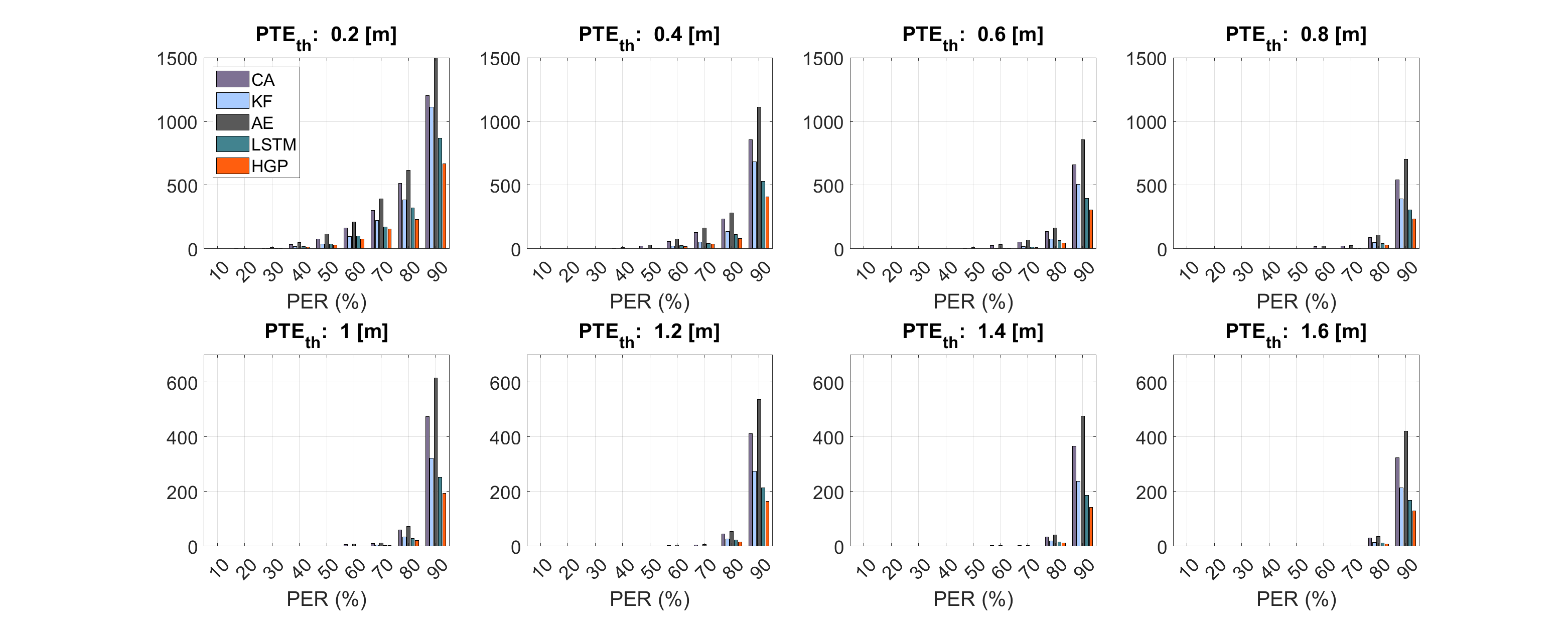}
\caption{Number of time which $PTE$ threshold has been exceeded vs $PER$ for different values of threshold ($PTE_{th}$)}
\label{fig:count_PER}
\end{figure*}

%% NEW REVIEW
Following in Table~\ref{table:compare} we present the tabular results for the performance of the baselines and our HGP approach in the SPMD dataset, measured by the $PTE$ at different $PER$ values. For completeness, we include both direct (AE-D, LSTM-D, HGP-D) and indirect (AE, LSTM, HGP) approaches. The baseline methods are CA, KF, AE-D, LSTM-D, HGP-D, AE, LSTM, and HGP. The results from Table~\ref{table:compare} show the improved performance when using the HGP in terms of $PTE$, which are more noticeable as communication losses increase measured by $PER$, verifying our \textbf{H3}.

\begin{table*}[!ht]
    \centering
    \caption{\small{Tabular results for the performance of the baselines and our HGP approach in the SPMD dataset, measured by the $PTE[m]$ at different $PER[\%]$ values. For both direct (AE-D, LSTM-D, HGP-D) and indirect (AE, LSTM, HGP) approaches.}}
    \begin{tabular}{l|l|l|l|l|l|l|l|l|l|l|l|}
    % \hline
        &  \multicolumn{11}{c}{PER [\%]}  \\ 
        \hline 
        \hline
        Methods & 0 & 10 & 20 & 30 & 40 & 50 & 60 & 70 & 80 & 90 & 95  \\ 
        \hline
        CA & 0.000 & 0.047 & 0.077 & 0.091 & 0.110 & 0.303 & 0.605 & 1.059 & 1.392 & 3.026 & 5.296  \\ \hline
        KF & 0.000 & 0.059 & 0.093 & 0.098 & 0.126 & 0.293 & 0.514 & 0.895 & 1.189 & 2.778 & 4.687  \\ \hline
        \hline
        AE-D & 0.000 & 0.127 & 0.322 & 0.541 & 0.774 & 1.154 & 1.813 & 2.200 & 2.640 & 5.926 & 9.558  \\ \hline
        LSTM-D & 0.000 & 0.110 & 0.292 & 0.356 & 0.559 & 0.683 & 0.865 & 1.417 & 1.781 & 4.231 & 6.918  \\ \hline
        HGP-D& 0.000 & 0.108 & 0.299 & 0.322 & 0.413 & 0.546 & 0.824 & 1.106 & 1.599 & 4.167 & 6.473  \\ \hline
        \hline
        AE & 0.000 & 0.065 & 0.103 & 0.108 & 0.138 & 0.523 & 0.866 & 1.185 & 1.508 & 4.156 & 6.156  \\ \hline
        LSTM & 0.000 & 0.051 & 0.085 & 0.100 & 0.122 & 0.243 & 0.366 & 0.595 & 0.731 & 2.329 & 3.825  \\ \hline
        HGP & 0.000 & 0.047 & 0.076 & 0.089 & 0.111 & 0.216 & 0.281 & 0.476 & 0.649 & 2.163 & 3.245  \\ \hline
        % ~ & ~ & ~ & ~ & ~ & ~ & ~ & ~ & ~ & ~ & ~ &   \\ \hline
         \hline
         \hline
  
    \end{tabular}
     \label{table:compare}
\end{table*}

%   \begin{figure}[h!]
%  \centering
%  \includegraphics[width=.48\textwidth]{Figsold/gp1.eps}
%      \caption{PTE vs PER for the SPMD dataset}
%      \label{Figure:SPMD1}
%  \end{figure}

%   \begin{figure}[h!]
%  \centering
%  \includegraphics[width=.48\textwidth]{Figsold/gp2.eps}
%      \caption{PTE vs Rate for the SPMD dataset}
%      \label{Figure:SPMD2}
%  \end{figure}

%% NEW REVIEW

% Following we select a new set of 50 totally random trajectories from the same SPMD data set. 
We perform a sensitivity analysis and compare the performance of CA, KF, AE, and LSTM against the proposed method based on $PTE$ and the number of times in which a specific $PTE$ threshold ($PTE_{th}$) has been exceeded. We investigate different $PTE_{th}$s, i.e. from 0.2 m to 1.6m. 
% These values cover the range between minimum and maximum thresholds specified by SAE J2945/1 and other extreme values. 
In general, vehicular safety applications are sensitive to position-tracking error in a hard-thresholding manner. The value of this threshold is dependent on the application and the driving scenario. The position error exceeding the $PTE_{th}$ can cause false positives and negatives in the safety applications and directly influence their performances. 

In this test, in the case of a packet loss, the HV performs prediction and if the prediction error is larger than the threshold, it is counted as an inadmissible prediction, possibly causing a hazardous situation (PTE higher that the admissible threshold $PTE_{th}$). In Figure \ref{fig:count_PER}, the total number of inadmissible predictions during the trajectories for different $PTE_{th}$s are plotted. % Finally we show the number of time that different threshold has been exceeded when PER increase in the communication. Figure \ref{fig:count_PER} shows the results.
For example, the plot in the top left shows the number of times that the $PTE$ is bigger than $PTE_{th}=0.2m$, which gives a sense of how many times an application that requires a $PTE$ less than $PTE_{th}=0.2m$ will fail. The bottom right shows the number of times that the $PTE$ is larger than $PTE_{th}=1.6m$ for different PERs. it shows that for a low $PER$ the $PTE$ never exceeds 1.6m, and for higher values of $PER$, the HGP performs better in all the cases. For instance, at $PER=90\%$ and $PTE_{th} = 1.6m$, CA fails more than $200\%$ times compared to the proposed HGP, similarly, the proposed HGP outperforms the other baselines. These results confirm the notable superiority of our approach as $PER$ increases and further validate hypothesis \textbf{H3}.
% Figure \ref{fig:count_Rate} shows the results for the number of times that a threshold has been exceeded for different values of $Rates$, in all the cases our approach outperforms CA, and in many the those cases improve the results by an amount bigger than 50 $\%$.

% First we compare the $PTE$ vs PER , and similar to previous results , our proposed has a superior performs , which is independent from the date selected. Figure \ref{fig:PER_PtcPTE} shows the results, the $PTE$ at PER 90 $\%$ for GP outperforms CA by 1m, which is a considerable amount.

% \begin{figure}[h]
% \centering
%  \includegraphics[width=.48\textwidth]{Figsold/Fig1PERv2.eps}
% % \includegraphics[width=10cm]{figure1.eps}
% \caption{ $PTE$ vs PER for the SPMD dataset, in a random selection of 50 trajectories}
% \label{fig:PER_PtcPTE}
% \end{figure}

% \begin{figure}[h]
% \centering
%  \includegraphics[width=.48\textwidth]{Fig1Rate.eps}
% % \includegraphics[width=10cm]{figure1.eps}
% \caption{ Rate vs 95 Ptc PTE}
% \label{fig:Rate_PtcPTE}
% \end{figure}

% \begin{figure}[h]
% \centering
%  \includegraphics[width=.48\textwidth]{Figsold/Fig3Rate.eps}
% % \includegraphics[width=10cm]{figure1.eps}
% \caption{ Number of time which threshold has been exceeded vs Rate for different values of threshold}
% \label{fig:count_Rate}
% \end{figure}

\section{Conclusion and Future Work}
In this study, we proposed a Context-Aware Target Classification module with a hybrid Gaussian Process regression as the prediction module to improve the robustness and flexibility of CVS systems. Unlike previous works that are strictly information message dependent, the proposed CA-TC allows safety applications to work based on the context-aware map rather than individual information messages, alleviating the effect of information loss on the performance of safety applications. The HGP relies on a reduced-size bank of driving models learned offline from real data to perform forecasting on-the-fly, while also allowing the online adding of new driving models to the bank, accounting for unforeseen driving behaviors.
We studied the impact of communication uncertainties, in terms of $PER$ and $Rate$, on the performance of the safety applications using the proposed CA-TC and compare it to the baseline. The performance is measured both in terms of $PTE$ and $WA$. The architecture is validated by simulation and real driving scenarios using three different datasets and a RV emulator based on a DSRC WSU. A notable performance improvement is observed using the proposed framework against traditional approaches.

% we demonstrate that our proposed architecture yields better performance than the previous approaches. 
% This is a very important and crucial finding which shows the strength of our proposed framework and makes it highly appealing for a realistic communication protocol design
% The proposed system is validated with comprehensive simulations and in a real environment using a Remote Vehicle Emulator (RVE) based and a DSRC WSU, which allows the joint study of the CVS applications and its underlying communication system. Additionally, we study the impact of using different prediction techniques and demonstrate the performance improvement of our proposed prediction method.
\smallskip
% \noindent 
\textbf{Limitations and Future Work. } % State any limitations
While we proposed a novel prediction method to be used in our architecture, the CA-TC can be further improved by the integration of other prediction methods. Data fusion and cooperative perceptions are among the possible methods to be incorporated with the flexible CA-TC architecture to further enhance the safety application's performance.
% For example, a machine learning-based technique can be used to choose between alternative motion models while taking driver behavior into consideration. 
In future works, we will examine the integration of error-driven model-based communication into our system. We plan to investigate the existence of natural and meaningful driving patterns for long-term maneuver and driver's intention prediction. We also plan to experiment with more CVS applications and other baselines. In a parallel research direction, we plan to extend our work to non-cooperative applications and test our predictive model in different tasks from commonly used driving datasets such as KITTI and Waymo, among others. 
\balance

\bibliography{main.bib}{}
\bibliographystyle{unsrt}
\end{document}